# Computer Vision-Based Model for Detecting Turning Lane Features on Florida's Public Roadways from Aerial Images

**Richard Boadu Antwi**
Graduate Research Assistant
Department of Civil and Environmental Engineering, Florida A&M University–Florida State University
College of Engineering, Florida State University
2525 Pottsdamer Street, Tallahassee, FL 32310, USA
**Corresponding author**, Email: rantwi@fsu.edu

**Samuel Takyi**
Graduate Research Assistant
Department of Civil and Environmental Engineering, Florida A&M University–Florida State University
College of Engineering, Florida State University
2525 Pottsdamer Street, Tallahassee, FL 32310, USA

**Kimollo Michael**
Graduate Assistant
School of Engineering, University of North Florida
Jacksonville, FL 32224

**Alican Karaer**
Associate Data Scientist
Iteris, Inc.
Tallahassee, FL, 32304, USA

**Eren Erman Ozguven**
Associate Professor
Department of Civil and Environmental Engineering, Florida A&M University–Florida State University
College of Engineering, Florida State University
2525 Pottsdamer Street, Tallahassee, FL 32310, USA

**Ren Moses**
Professor
Department of Civil and Environmental Engineering, Florida A&M University–Florida State University
College of Engineering, Florida State University
2525 Pottsdamer Street, Tallahassee, FL 32310, USA

**Maxim A. Dulebenets**
Associate Professor
Department of Civil and Environmental Engineering, Florida A&M University–Florida State University
College of Engineering, Florida State University
2525 Pottsdamer Street, Tallahassee, FL 32310, USA

**Thobias Sando**
Professor
School of Engineering, University of North Florida
Jacksonville, FL 32224

*Word Count 8,766*

## ABSTRACT

Efficient and current roadway geometry data collection is a critical task for transportation agencies to undertake effective road planning, maintenance, design, and rehabilitation efforts. The methods for gathering such data can be broadly classified into two categories: a) land-based methods, which encompass field inventory, mobile mapping, and image logging, and b) aerial-based methods, which involve satellite imagery, drones, and laser scanning. However, employing land-based techniques for extensive highway networks covering thousands of miles proves arduous and costly, and poses safety risks for crew members. Consequently, there exists a pressing need to develop more efficient methodologies for acquiring this data promptly, safely, and economically. Fortunately, with the increasing availability of high-resolution images and recent strides in computer vision and object detection technologies, automated extraction of roadway geometry features has become feasible. Proposing a computer vision-based methodology, this novel study aims to detect turning lane pavement markings from high-resolution aerial images and extract turning lanes on Florida's public roadways. This endeavor holds paramount importance for transportation agencies, serving various purposes such as identifying aged or faded markings, comparing turning lane locations with other geometric features like crosswalks, and analyzing crashes occurring around intersections. In comparison with ground truth data obtained for Leon County, Florida, the developed model observed an average accuracy of 80.4% at the lowest confidence threshold for the detected features. The model was then used to detect turning lane markings in Duval County, Florida and approximately 8,737 left turn, 3,577 right turn, and 1,520 center lane features were identified automatically. The roadway geometry data extracted can be seamlessly integrated with crash and traffic data, offering invaluable insights to policy makers and roadway users.

## INTRODUCTION

The progression of computer vision technology is swiftly enabling traffic agencies to economize time and resources in the collection of roadway geometry data. A 2015 study (1) revealed that, gathering geometry data using aerial and satellite imagery proved more favorable than traditional field observations regarding equipment expenses, data precision, crew welfare, data collection expenses, and collection duration. Historically, image processing has been acknowledged as a labor-intensive and fallible technique for capturing roadway data. However, advancements in computational capabilities and image recognition techniques have created fresh opportunities for accurately detecting and mapping various roadway attributes from a variety of imagery sources.

Following the introduction of the Highway Safety Manual (HSM) in 2010, numerous state departments of transportation (DOTs) have prioritized safety enhancement by integrating the manual's geometric data prerequisites into their data collection endeavors (2). Nonetheless, manually gathering data across extensive roadway networks has posed a significant challenge for many state and local transportation agencies. Central to the responsibilities of these agencies is the upkeep of current roadway data, which is crucial for effective roadway planning, maintenance, design, and rehabilitation efforts (3). Roadway characteristics index (RCI) data encompass a comprehensive inventory of all elements constituting a roadway. This includes various facets such as Highway Performance Monitoring System (HPMS) data, roadway geometry, traffic signals, lane counts, traffic monitoring locations, turning restrictions, intersections, interchanges, rest areas with or without amenities, High Occupancy Vehicle (HOV) lanes, pavement markings, signage, pavement condition, driveways, and bridges. DOTs employ different methodologies for RCI data collection, ranging from field inventory and satellite imagery to mobile and airborne Light Detection and Ranging (LiDAR), integrated Geographic Information System (GIS)/Global Positioning System (GPS) mapping systems, static terrestrial laser scanning, and photo/video logging (2).

These techniques can be categorized as ground observation/field surveying and aerial imagery/photogrammetry. The process of collecting RCI data incurs significant costs. Each approach necessitates distinct equipment and time allocation for data gathering, thereby impacting the overall expense. In ground surveying, measurements are conducted along the roadside using total stations. Field inventory entails traversing the roadway to document current conditions, subsequently inputting this information into the inventory database. Conversely, aerial imagery or photogrammetry involves extracting RCI data from georeferenced images captured by airborne systems such as satellites, aircraft, or drones. Each approach presents its own set of pros and cons concerning data acquisition costs, data accuracy, quality of data retrievable, constraints on data collection, storage requirements, labor intensity, acquisition and processing duration, and crew safety. Moreover, pavement markings typically last between 0.5 and 3 years (4). Due to their short lifespan, frequent inspection and maintenance have been necessary. A periodic manual inspection approach is generally adopted by road inspectors to evaluate pavement marking conditions, which is time-consuming and risky.

The majority of highway agencies and DOTs traditionally gather roadway data through direct field observations, despite the inherent challenges of these methods, which are often difficult, time-consuming, pose risks and susceptible to limitations, particularly in adverse weather conditions (3). Hence, there is a pressing need to investigate alternative and more efficient approaches for gathering roadway inventory data. Researchers have delved into novel and emerging technologies for the acquisition of roadway inventory data, including image processing, and computer vision methods. Over the past decade, satellite and aerial imagery have been leveraged extensively to gather earth-related information (2). High-resolution images captured from aircraft or satellites can swiftly yield RCI data upon processing (5-7). In a previous study, AI was used to extract school zones high-resolution aerial images (7). The growing availability of such images, coupled with advancements in data extraction techniques, underscores the imperative to optimize their utilization for extracting roadway inventory data efficiently. While a challenge lies in extracting obscure or small objects, recent progress in machine learning research has mitigated this constraint. Well-trained models can enhance output accuracy significantly. Moreover, this method incurs minimal to no field costs, rendering it economically favorable.

RCI data also include essential details such as such as center, left, or right turning lanes, which are critical information required by State DOTs as well as local transportation agencies. These pavement markings, mostly close to intersections, indicate which turn may be made from a lane to guide and control traffic for roadway users. There are several benefits of these roadway markings expressed in a USDOT study (8). For instance, roadway markings offer ongoing guidance on vehicle positioning, lane navigation, and roadway alignment. Several DOTs have shown major interest in automated pavement marking detection and assessment methods (9). Turning lane data plays a crucial role in enhancing roadway efficiency and reducing crashes, especially at intersections. However, a comprehensive geospatial data inventory of turning lanes is lacking, focusing on both local and state roadways in Florida. Hence, it becomes imperative to devise innovative, rapid, and effective methods for gathering such information. This holds significant importance for transportation agencies, serving various purposes such as identifying old or obscured markings, juxtaposing turning lane positions with other geometric features like crosswalks and school zones, and analyzing accidents occurring in these zones as well as at intersections.

To the authors’ knowledge, there has not been any research into utilizing computer vision techniques for extracting and creating an inventory of turning lane markings using high-resolution aerial images in the literature. Consequently, this study aims to explore the development of automated tools for detecting these roadway features through deep learning-based object detection models. Specifically, the study will concentrate on creating an automated You Only Look Once (YOLO)-based artificial intelligence model to identify and extract turning lane markings such as left, right, and center from high-resolution aerial images. This will involve the development of two distinct multiclass object detection models to compare their effectiveness, followed by the application of the chosen model to detect the left, right, and center lanes throughout the State of Florida. This endeavor is crucial for transportation agencies for several reasons, including identifying aging or obscured markings, comparing turning lane positions with other geometric features such as crosswalks and school zones, and analyzing accidents occurring within these zones and at intersections. Furthermore, the automatically extracted road geometry data can be integrated with crash and traffic data to provide insights to policymakers and roadway users.

# LITERATURE REVIEW

Recently, there has been a surge in research concerning the extraction of roadway RCI data, utilizing artificial intelligence (AI) methods, particularly pertaining to pavement markings. AI techniques facilitate the extraction of roadway information from such imagery through the application of detection models. One example is (7), which employed YOLO to identify school zones from high-resolution aerial images. Other research endeavors focusing on RCI data collection methods have predominantly emphasized the utilization of LiDAR technology, which generates comprehensive data through point clouds. However, this approach is hindered by its high equipment and operational costs, challenges in processing large datasets, and lengthy data processing and reduction durations. For instance, a study (3) explored the use of LiDAR for creating inventories of various roadway features, such as highway grade, side slope, contours, and stopping and passing sight distance, employing high-speed computing, GPS laser range finders, and precision inertial navigation for data collection. Additionally, Mobile Terrestrial Laser Scanning (MTLS) has been employed to extract highway inventory data in other research endeavors (5). It is worth noting that other studies in the field of autonomous driving have recently employed computer vision, sensor technology, and AI techniques to identify pavement markings for vehicle navigation (10-14).

Other studies have concentrated on conventional methods, characterized by prolonged data collection periods, disruption of traffic operations, and exposure of crew members to various hazards. Emerging technologies such as deep learning, object detection, and computer vision skills have received scant attention in comparison. These technologies enable the extraction of roadway features from high-resolution aerial images, effectively eliminating the need for fieldwork and reducing data collection time. Moreover, the data obtained through these methods is readily accessible, poses no risks to crew members in terms of traffic-related dangers with high accuracy, and swiftly acquires roadway features across extensive areas. In selecting the most suitable object detection model, performance based solely on standardized image collection is not the sole criterion (15). Various factors must be considered due to the project's unique requirements. Notably, many top models, including those provided by TensorFlow, have converged on similar fundamental principles, facilitating their applicability for specific object identification tasks. Consequently, each model's performance is typically defined in terms of speed, memory usage, and accuracy.

### *Utilizing Computer Vision and Deep Learning for Roadway Geometry Feature Extraction*

Two emerging examples of computer vision and deep learning techniques gaining traction among transportation experts are convolutional neural networks (CNN) and recurrent neural networks (RNN) (16). Researchers have utilized CNN and R-CNN in roadway geometry data extraction, achieving notable advancements. These techniques have demonstrated the ability to efficiently recognize, detect, and map roadway geometry features across expansive areas in a short timeframe, requiring minimal human intervention. For instance, R-CNN was used in (4) to automatically detect roadway pavement marking defects from photogrammetric data obtained from Google Maps to assess pavement conditions. In this study, the created model was trained to identify defects such as wrong alignment, ghost marking, missing or faded edges, corner, or segments and cracks, in the pavement markings with an average accuracy of 30%.

In a study, computer vision algorithms were employed to identify and classify traffic signs using Google Street View (GSV) images captured along interstate segments (17). The researchers utilized a machine learning technique that combined Histograms of Oriented Gradients (HOG) features with a linear Support Vector Machine (SVM) classifier to categorize roadway signs based on their pattern and color information. They achieved a classification accuracy of 94.63%. Nevertheless, the method was not evaluated on local or non-interstate roadways. The study conducted by (18) employed a method called ROI-wise reverse reweighting network to identify roadway markings from ground-level images. By emphasizing multi-layer features, the study utilized multi-layer pooling operation and ROI-wise reverse reweighting techniques, based on Faster RCNN, achieving a detection accuracy of 69.95% for roadway markings in images.

In (12), CNN was applied to identify lane markings on roadways and estimate roadway geometries using inverse perspective mapped images. In (19), scholars utilized CNN to identify, quantify, and extract concealed cracks' geometric characteristics in roadways from ground-penetrating radar images. In (20), authors employed CNN to extract roadway features from remotely sensed images. These techniques have been applied for image detection for diverse objectives. For instance, autonomous vehicles employ them to detect objects and infrastructure on roadways in real-time. For experimentation, a novel hybrid local multiple system (LM-CNN-SVM) combining convolutional neural network (CNN) and SVMs was tested using the Caltech-101 and Caltech pedestrian datasets to detect objects and pedestrians (21). The method involved partitioning the entire image into local regions, with multiple CNNs tasked to learn characteristics of local areas and items. Principal component analysis was used to select significant features, which were then fed into multiple SVMs employing empirical and structural risk reduction techniques instead of direct CNN usage to enhance the classifier system's generalization capacity. Furthermore, both a pre-trained AlexNet and a unique CNN architecture were employed, with the SVM output being combined. The method demonstrated an accuracy range between 89.80% and 92.80%.

Recent advancements in object identification have showcased the remarkable capabilities of convolutional neural networks (CNN) based on deep learning (22–25). The evolution from region-based CNN (R-CNN) to fast R-CNN (26), and eventually faster R-CNN (27) has significantly improved these algorithms. Essentially, the CNN approach entails extracting region proposals as potential object locations and then computing CNN features. Faster R-CNN further enhances performance by incorporating an additional Region Proposal Network (RPN) into the object detection network. In a study (28), a comprehensive comparison was conducted between machine learning algorithms and deep learning methods, namely faster R-CNN and ACF, in terms of their vehicle recognition and tracking capabilities. The findings indicate that faster R-CNN outperforms ACF in this regard. Hence, careful consideration is warranted when selecting the algorithm for conducting the analysis. Depending on the objectives of the analysis, the output of vehicle identification and tracking algorithms may encompass parameters such as speed, volume, or vehicle trajectories. Additionally, vehicle detection algorithms have the capability to categorize cars, which distinguishes them from point tracking methods.

### *Obtaining Roadway Geometry Data through LiDAR and Aerial Imagery Techniques*

As road markings typically exhibit high reflectance in images, this property can be leveraged for their extraction from images. Researchers in (29) devised an algorithm reliant on pixel extraction to identify and detect road markings from images captured by cameras mounted on moving vehicles. Initially, the algorithm extracted marking pixels using a median local threshold (MLT) image filtering method. Subsequently, a recognition algorithm was employed to detect markings based on their shapes and dimensions. The average true positive rate for the detected markings stood at 84%. Researchers in (30) utilized a CNN-based semantic segmentation technique to extract lane markings from aerial images. The study introduced a novel approach employing a fully convolutional neural network (FCNN) and discrete wavelet transform (DWT) to process high-resolution images for lane marking extraction. Unlike previous methods, this approach, featuring a symmetric FCNN, specifically targeted small-sized lane markings and explored optimal DWT insertion points within the architecture. By integrating DWT into the FCNN architecture, the model could effectively capture high-frequency details, thereby improving lane detection accuracy. The proposed Aerial LaneNet architecture comprised a series of convolutional, wavelet transform, and pooling layers. Evaluation of the model's performance was conducted using a publicly available aerial lane detection dataset. The study noted limitations in areas with shadows and washed-out lane markings.

In (31), researchers applied the scan line technique to extract roadway markings from LiDAR point clouds. By sequencing the LiDAR point clouds based on their timestamps and organizing them into scan lines according to the scanner angle, the team effectively arranged the data. They then leveraged the disparity in height between trajectory data and the road surface to isolate seed roadway points, which served as the basis for deriving complete roadway points. These points were used to establish a line that encompassed the seed points and all other points along the scan line. Following this, points falling within a specified threshold of this line were retained and categorized as either road markings or asphalt points based on their intensity. To enhance data quality by reducing noise, a dynamic window median filter was applied to smooth out intensity values. Lastly, the research team utilized edge detection techniques and constraints to identify and extract roadway markings. Researchers in (3) used aerial imagery with a resolution of two meters and LiDAR data stored in ASCII comma-delimited text files with an accuracy of six inches to gather roadway data. They augmented LiDAR point shapefiles with height information using ArcView software and subsequently transformed the data points into Triangular Irregular Networks (TIN). The LiDAR bounds and aerial images were combined to initiate the analysis process. ArcView's object obstruction tool was employed to detect obstacles obstructing the observer's line of sight. Stopping sight distance was determined by tracing the line of sight, with visible terrain represented by green line segments and obstructed terrain by red line segments. Side slope and contours were derived using the identification and contour tools in ArcView. Additionally, the researchers calculated the grade by computing the difference in elevation between the two ends of a segment. On the other hand, (32) utilized a 3D profile of pavement data acquired through laser scanning to construct a deep learning model for the automated detection and extraction of roadway markings. They achieved a detection accuracy of 90.8%. The study involved creating a step-shaped operator to pinpoint potential areas containing edges of pavement markings. These regions' geometric attributes and convolution features were then merged to extract the road markings using CNN.

Computer vision techniques were employed in (33) to extract pavement markings from video imagery and evaluate pavement conditions. The study involved obtaining videos captured from the frontal perspective of vehicles, which were then subjected to preprocessing. This preprocessing included extracting image sequences, filtering and smoothing the images using the Gaussian blur median filter and rectifying them using the inverse perspective transform. Pavement markings were detected using a hybrid detector based on gradient characteristics and color. Following this, image segmentation was employed to isolate regions containing markings, which were subsequently classified into categories such as edge lines, dividing lines, barrier lines, and continuity lines.

To date, there has not been any research examining the utilization of image processing and computer vision methods to create a turning lane marking inventory. Consequently, this study aims to address this gap by developing automated tools for detecting these turning lane markings (e.g., left, right and center). Specifically, the study will focus on creating an artificial intelligence model based on the You Only Look Once (YOLO) approach to accomplish this task.

## STUDY AREA

Duval County has been selected as the case study area for detecting turning lanes and configurations (**Figure 1**). The City of Jacksonville is the county seat for Duval County, which is located in northeast Florida. The county has 918 square miles of total area and shares border with Clay, Baker, Nassau, and St. Johns Counties in Florida (34). According to (34), Duval County has a population of 995,560. Duval County has over 587 miles of local roadways and 503 miles of state roadways. The county was considered due to its diversity in roadway infrastructural development. Ground truth data obtained from Florida's Leon County was used for validating the models' performance. Leon County, which is the home to the capital city of Florida; Tallahassee, has a population of 292,198 and 668 square miles of total area (35). It shares bother with other Florida counties like Gadsden, Wakulla, Liberty, and Jefferson as well as Grady and Thomas counties in Georgia.

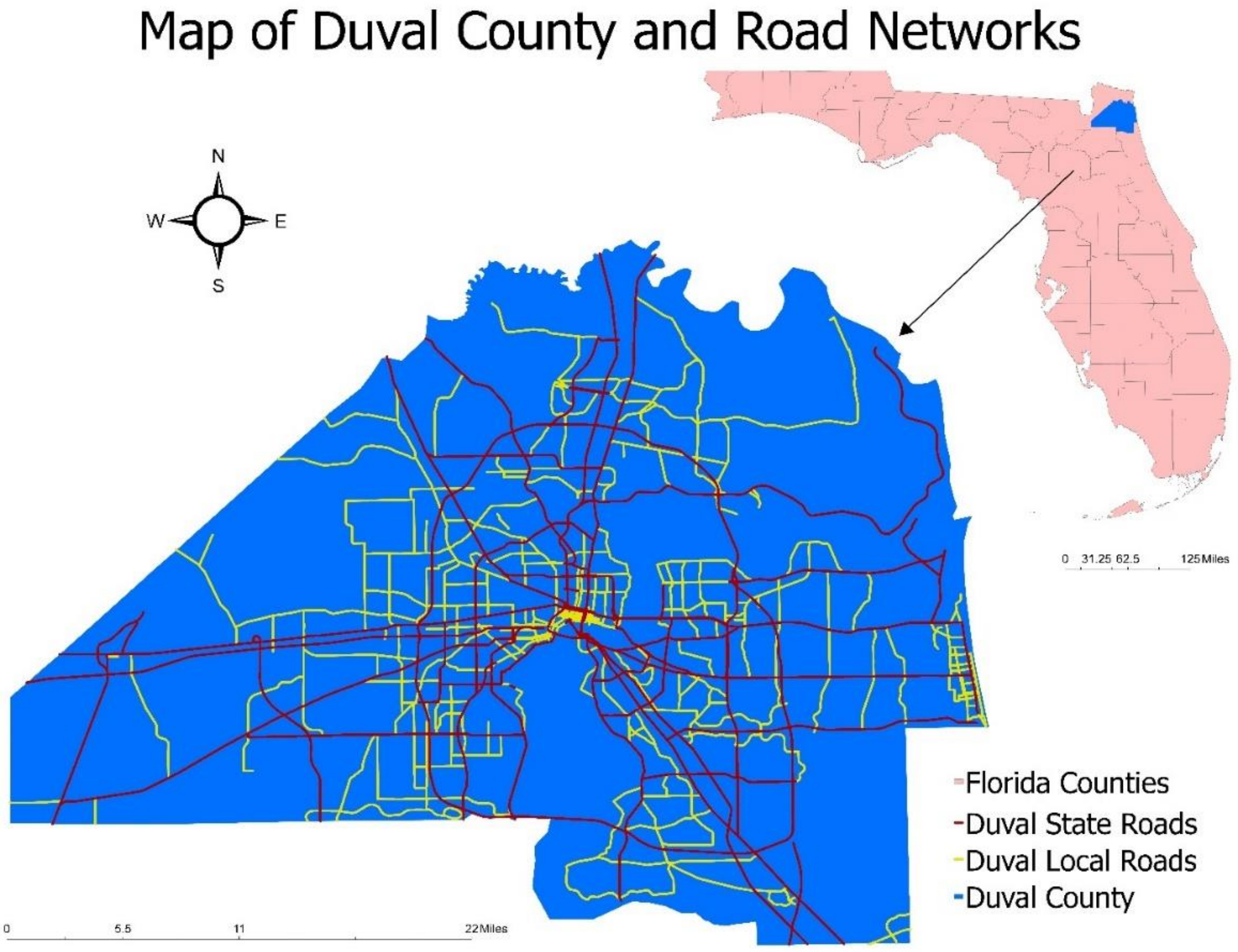

**Figure 1** Map of Duval County, Florida with the roadway network

# MATERIALS AND METHODOLOGY

The selection of the methodology for gathering roadway inventory data is contingent upon various factors including data collection time (e.g., data collection, reduction, and processing), cost (e.g., data collection, and reduction), as well as accuracy, safety, and data storage demands. In this study, our objective was to develop a deep learning object detection model tailored to identify turning lane markings from high-resolution aerial images within Duval County, Florida.

## *Data Description*

The proposed algorithm for detecting turning lanes, driven by artificial intelligence, concentrates on harnessing high-resolution aerial images alongside recent strides in computer vision and object detection methodologies. The objective is to streamline the process of extracting turning lane pavement markings, offering a novel solution for transportation agencies. The Surveying and Mapping Office of the Florida Department of Transportation (FDOT) maintains an archive of aerial imagery covering all 67 counties in Florida. These images are georeferenced and well-indexed, with filenames containing the county code, year of capture, and tile number. Access to this archive is facilitated through the Florida Aerial Photo Look-Up System (APLUS), operated by the FDOT Surveying and Mapping Office. While some photos can be downloaded from the APLUS website, acquiring large datasets requires a formal request or provision of external storage. For this study, high-resolution images from counties including Miami-Dade, Gulf, Santa Rosa, Hillsborough, Duval, Broward, and Leon were obtained, totaling 11.16 GB in size. These images, used for model training and detection, range in resolution from 1.5 ft down to 0.25 ft, allowing for effective detection of school zones. Most images have a resolution of 0.5 ft/pixel, with dimensions of 10,000 × 10,000 and a 3-band (RGB) format, though precise resolution varies by county. Images are provided in MrSID format, facilitating GIS projection on maps. Another data utilized are the state and local roadway shapefiles which were also obtained from FDOT’s GIS database.

This study concentrates on identifying turning lanes present on county- or city-managed roads, as well as those situated on state highway system routes. Interstate highways were excluded from the state road dataset, and all centerlines from the county and city-managed road shapefile were merged. According to FDOT categorization, state highway system routes are termed ON System Roadways or state roads, while county- or city-managed roads are referred to as OFF System Roadways or local roads. It is important to highlight that while FDOT's GIS data can provide various geometric data points crucial for mobility and safety assessments, it lacks information on the locations of turning configurations on state and local roads. Hence, the primary aim of this project is to utilize an advanced object identification model to compile an inventory of turning lane markers—specifically left-only, right-only, and center lanes—on both state and local roads in Duval County, Florida.

## *Pre-processing*

Due to the volume of data and the complexity of object recognition, preprocessing is essential. Our method involves selecting and discarding images that do not intersect with a roadway centerline and masking out pixels outside a buffer zone. This process reduces the number of photos from 90K to 30K and excludes objects beyond 100 feet from state and local roadway centerlines. Prior to masking the images, the roadway shapefile is buffered to create overlapping polygons, serving as reference for cropping intersecting regions of aerial images. During masking, pixels outside the reference layer's boundary are removed. The resulting cropped images, with fewer pixels, are then mosaiced into a single raster file, facilitating easier handling and analysis. **Figure 2** details this preprocessing approach.

Initially, all photos from selected counties are imported into a mosaic dataset using ArcGIS Pro. Geocoded photos are managed and displayed using mosaic databases, enabling intersection with additional vector data to select picture tiles based on location. Subsequently, a subset of photos (30K) is extracted, consisting of images intersecting with roadway centerlines. An automated picture masking tool is developed using ArcGIS Pro's ModelBuilder interface. This tool iterates through a photo folder, applies a mask based on a 100-foot buffer around road centerlines, and saves the masked images as JPG files for object recognition.

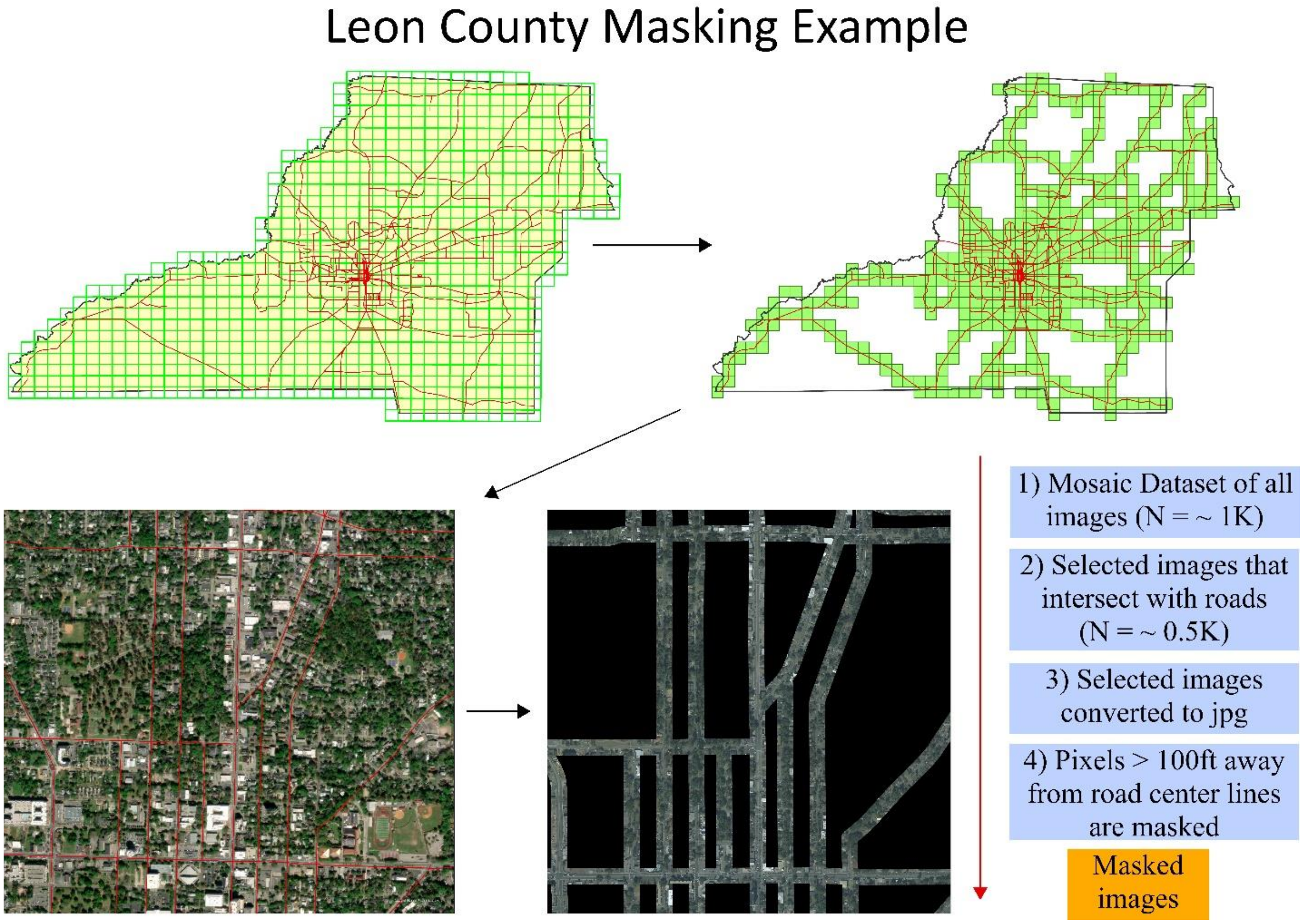

**Figure 2** Preprocessing approach, and automated image masking model used in the fourth step of preprocessing.

### *Data Preparation for Model Training and Evaluation*

High-quality image data is crucial for training deep learning models to identify various turning lane features. The process of image preparation, depicted in **Figure 3a**, outlines the steps from turning lane identification to the extraction of training data or images. Significant time and effort were dedicated to generating training data for model creation, as a substantial portion of the model's performance hinges on the quantity, quality, and diversity of the training data used. Similarly, additional time and effort were invested in the model training process, involving testing different parameters, and selecting the optimal ones to train the model. This is depicted by the larger box sizes for "Manually label turning lane features" and "training data" in **Figure 3a**.

For the initial investigation, two distinct multi-class models were developed to identify the turning configurations. Initially, a 12-class object detection model was created, followed by a 4-class object detection model. Different sets of training data with varying classes were prepared for each model. The first training data used for the initial turning lane model comprised 12 classes: "left_only," "right_only," "left_through," "right_through," "through," "left_right_through," "bicycle," "center," "left_right," "merge," "u_turn," and "none," denoted as classes 1 through 12 respectively. On the other hand, the second model's training data consisted of "left_only," "right_only," "center," and "none," corresponding to classes 1, 2, 8, and 12 respectively (**Table 1**). These labels were represented by rectangular bounding boxes encompassing the turning lane markings (**Figure 3b**). It is worth noting that object detection models perform optimally when trained on clear and distinctive features. Given that left-only and right-only turning markings often exhibit a lateral inversion of one another, rendering them less distinguishable, the model's performance in detecting them is expected to be lower. To ensure uniqueness, the labels for training the center lane class featured both left arrows facing each other from different travel directions. This approach was similarly applied to label all other features to maintain uniqueness in the training features. The metadata of the exported labels followed the Pascal Visual Objects format. The input mosaic data comprised high-resolution aerial images covering the entire State of Florida, with a tile size of 5000x5000 square feet.

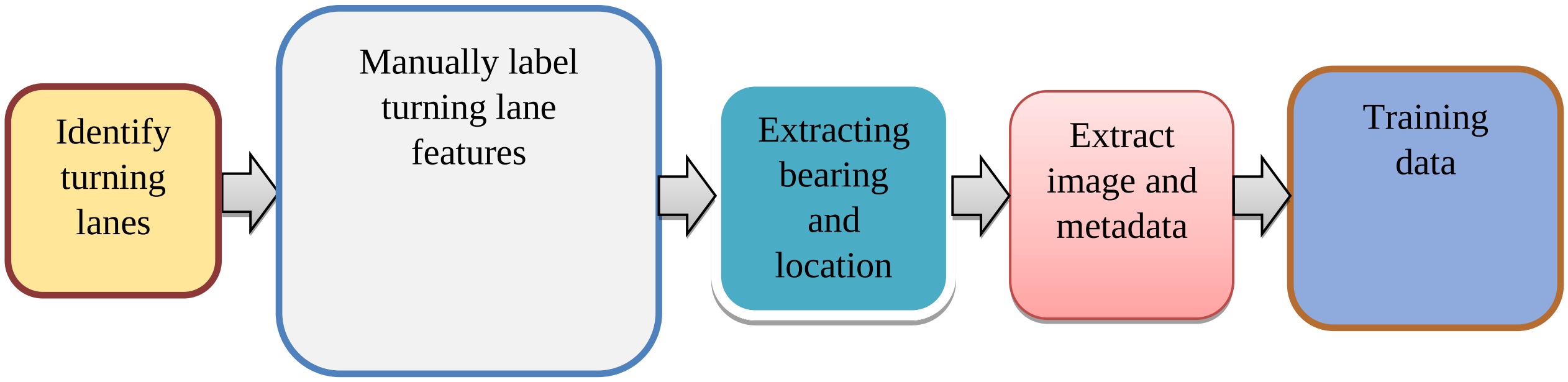

**(3a)**

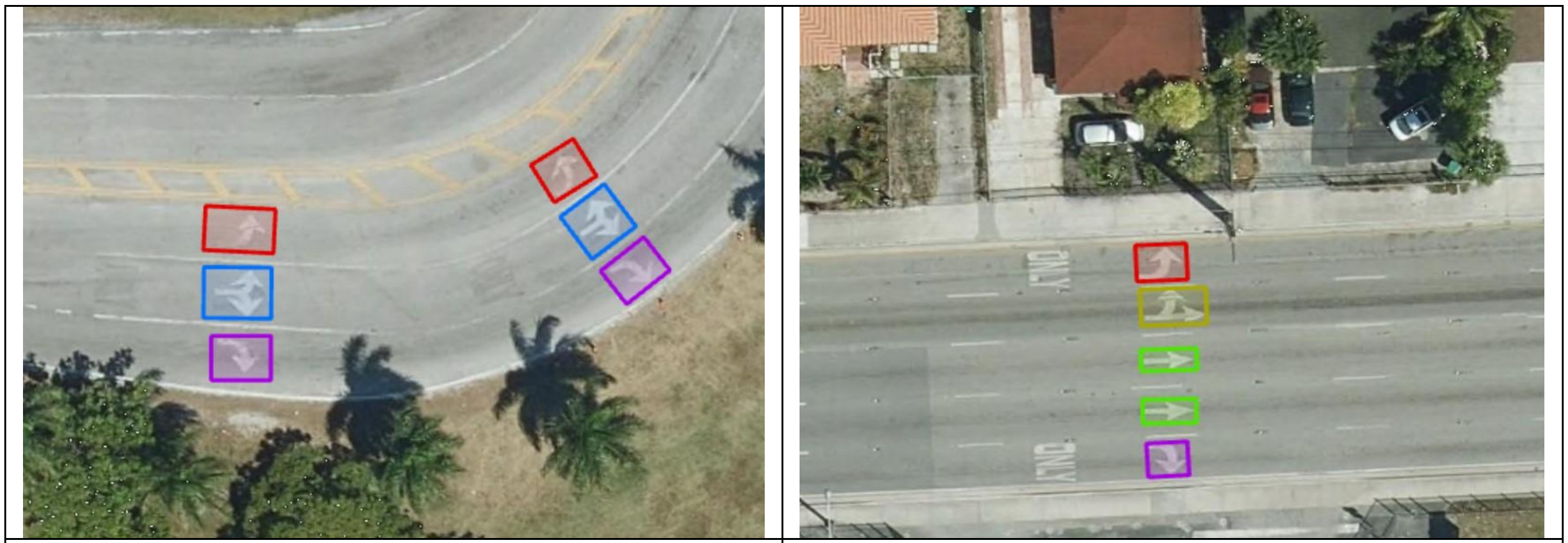

**(i) left_only, left_right, and right_only**

**(ii) left_only, left_through, through, and right_only**

| | |
|---|---|
| 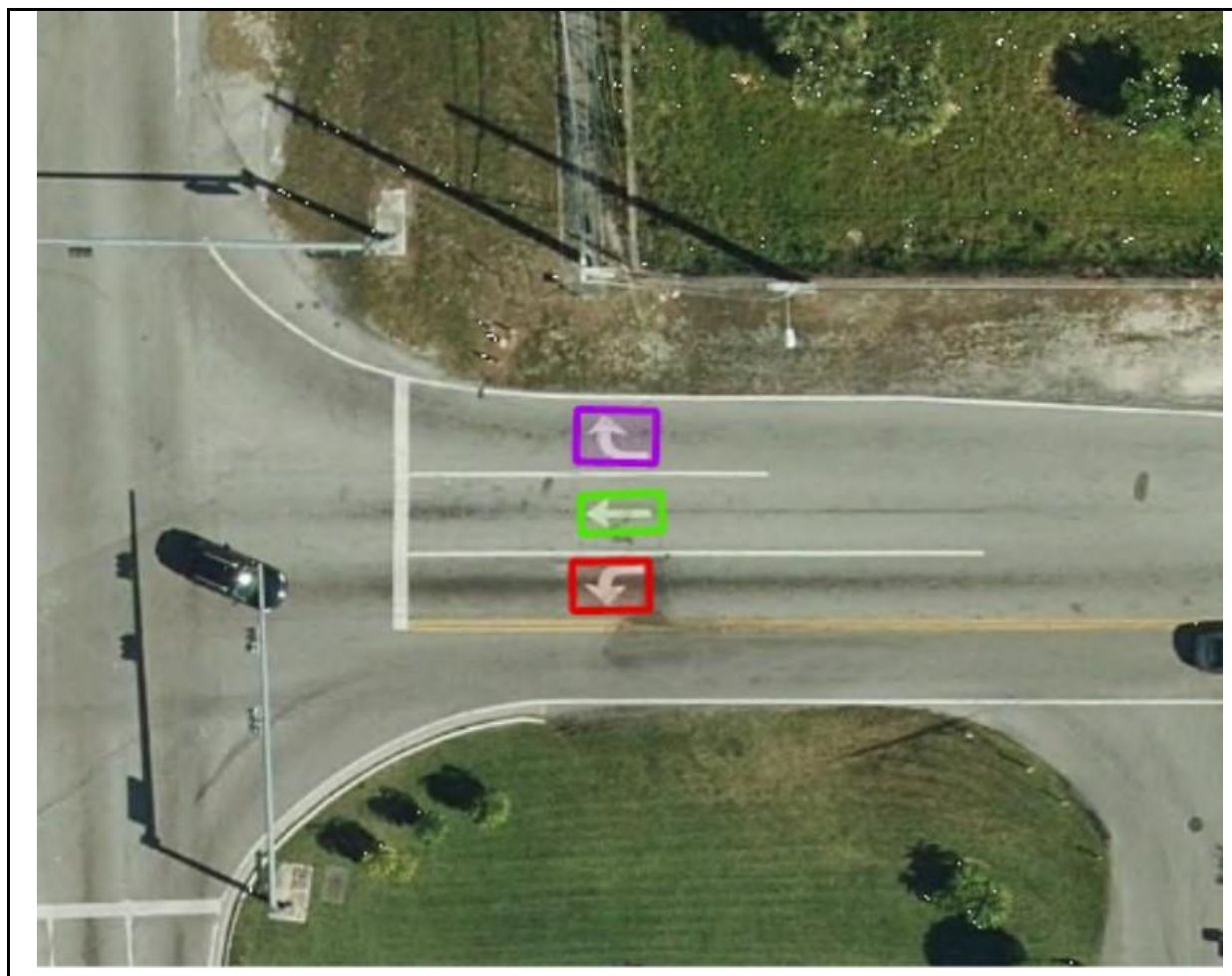 | 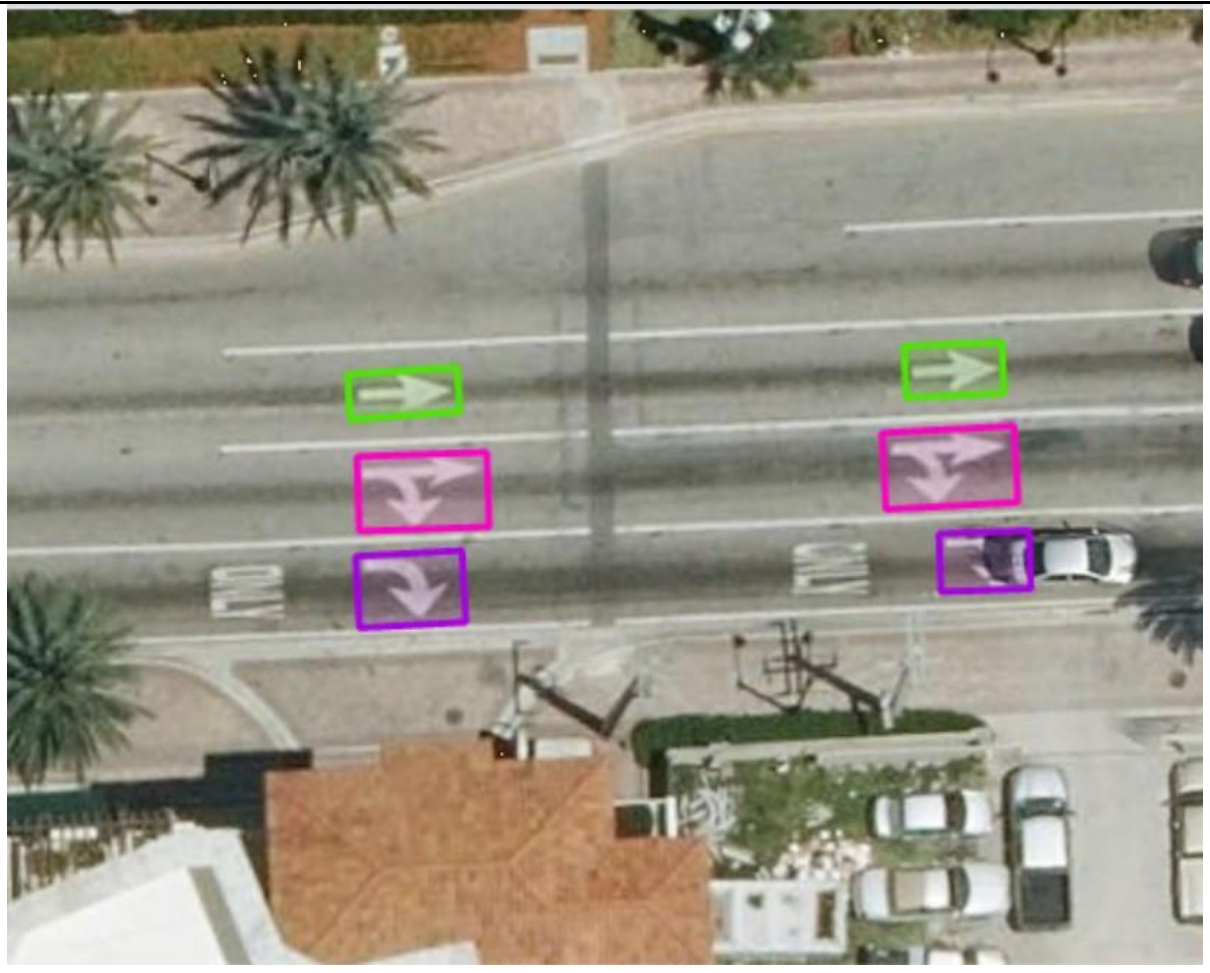 |
| **(iii) right_only, through, and left_only** | **(iv) through, right_through, and right_only** |

**(3b)**

**Figure 3** (a) Model training data preparation framework, and (b) training data examples with bounding boxes.

**Table 1**: Training data description

| **ID** | **Class name** | **Description** | **Example/Picture** | **Training Example** |
|---|---|---|---|---|
| 1 | left_only | Left only | 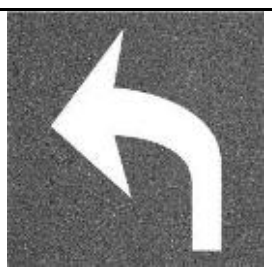 | 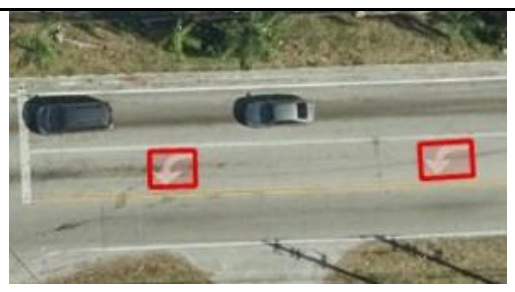 |
| 2 | right_only | Right only | 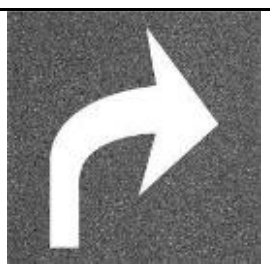 | 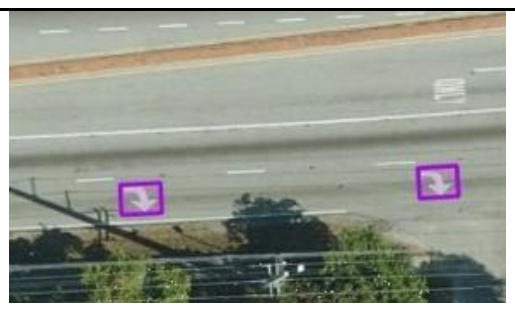 |
| 3 | left_right | Left and Right | 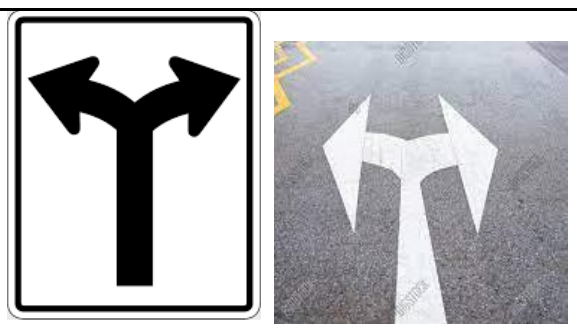 | 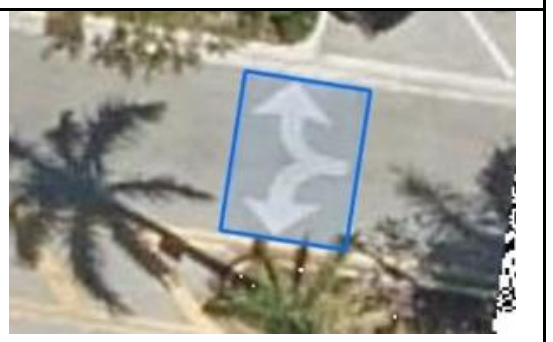 |
| 4 | through | Through | 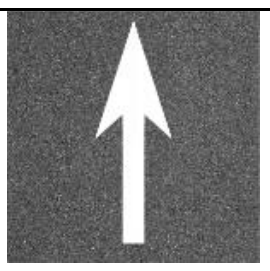 | 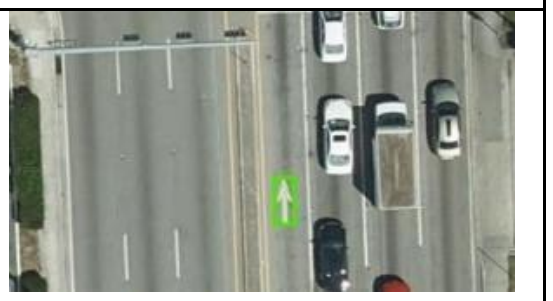 |
| 5 | left_through | Left and Through | 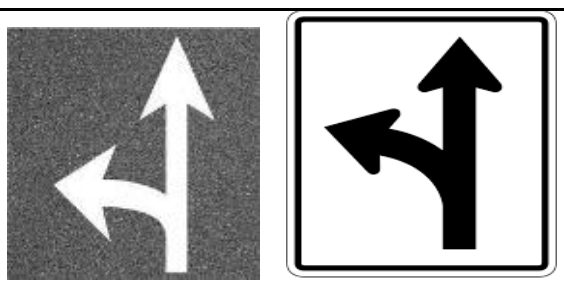 | 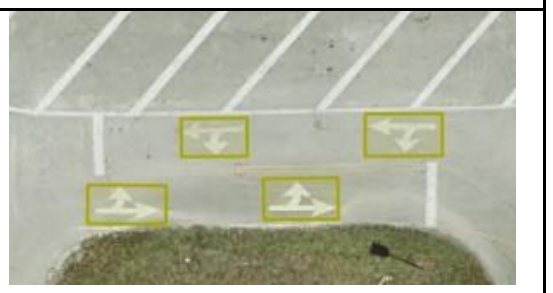 |
| 6 | right_through | Right and Through | 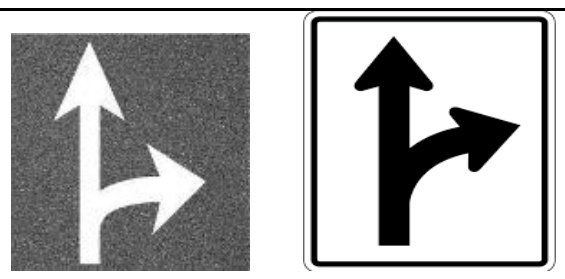 | 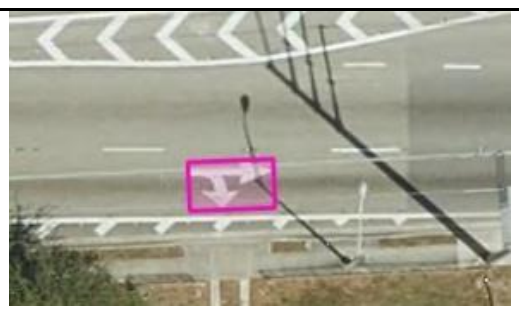 |
| 7 | left_right_through | Left Right and Through | 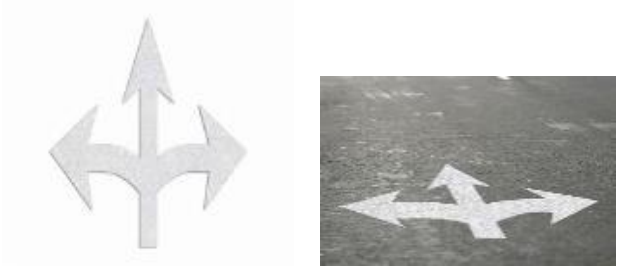 | 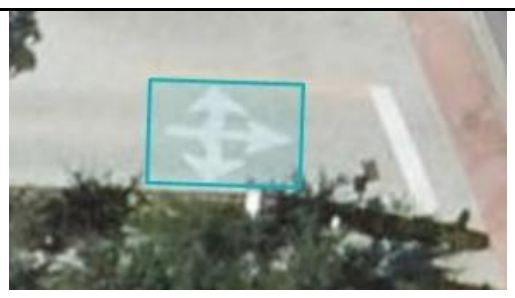 |
| 8 | center | Center lane (left turn possible in both direction or two way) | 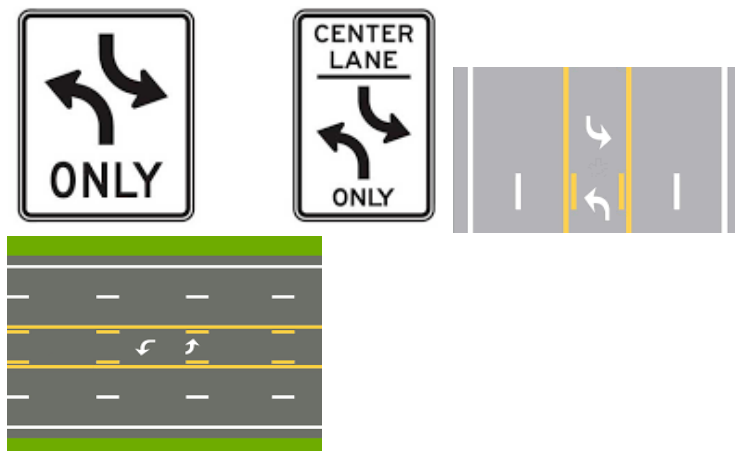 | 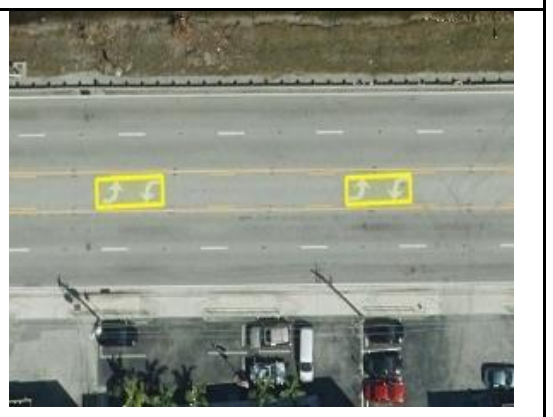 |
| 9 | bicycle | Bicycle lane | 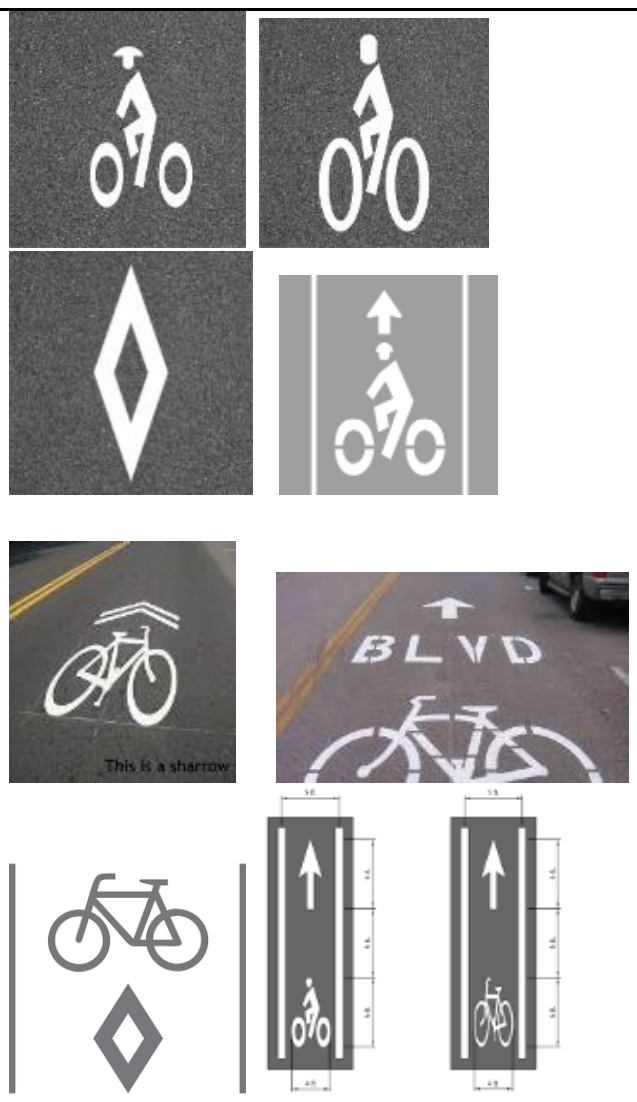 | 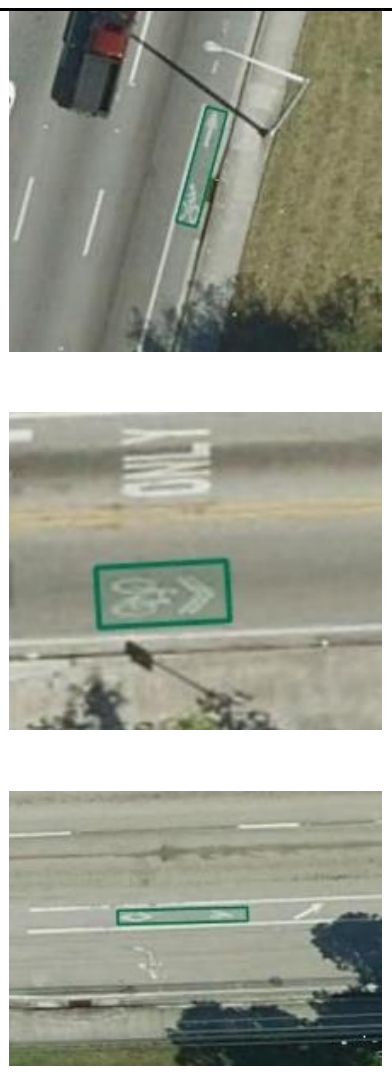 |
| 10 | merge | Merge lane | 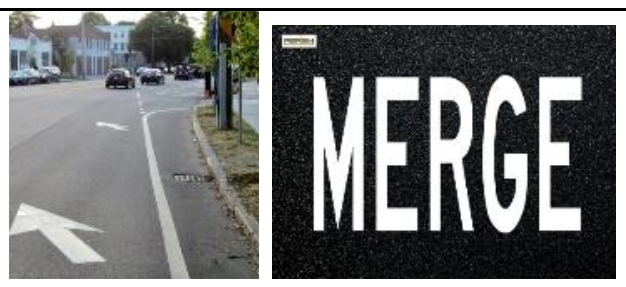 | 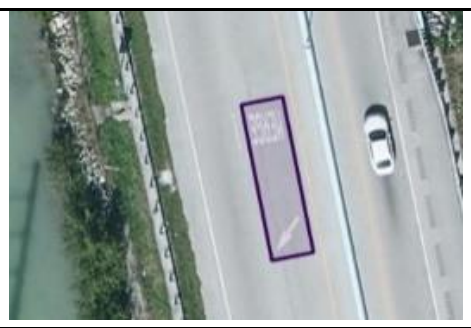 |
| 11 | u_turn | U turn | 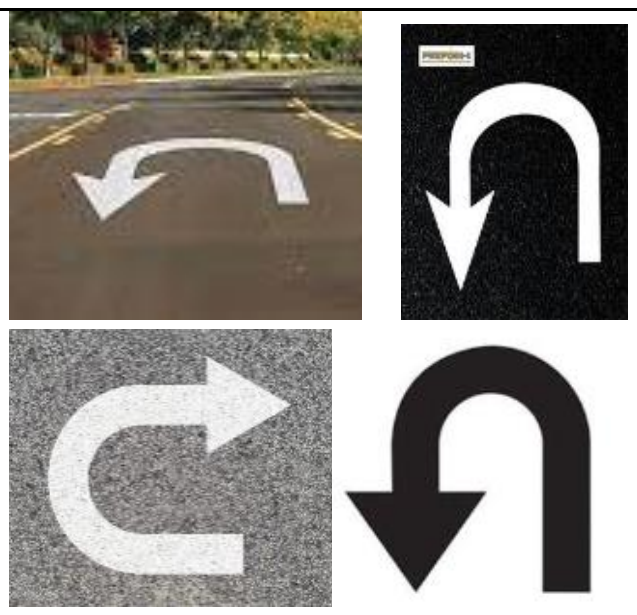 | 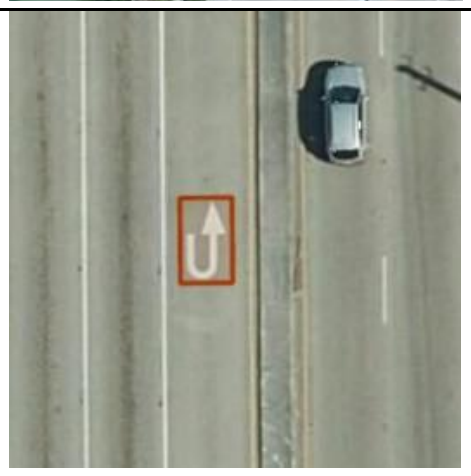 |
| 12 | none | None<br>1. Yield<br>2. Stop<br>3. Parking<br>4. Speed limits | 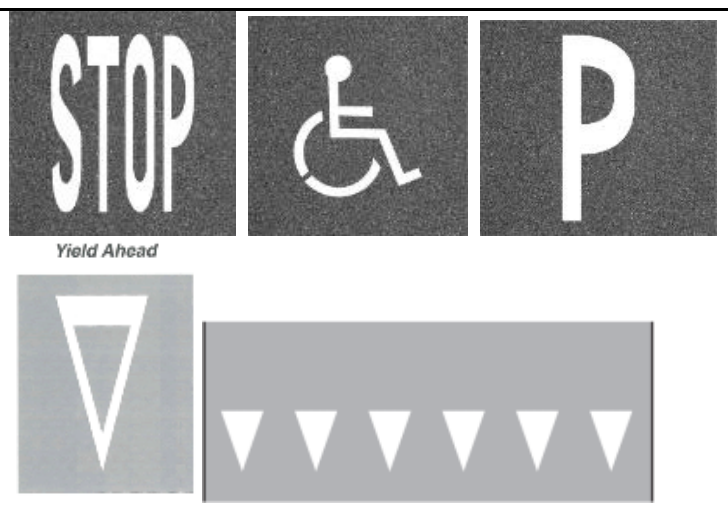 | 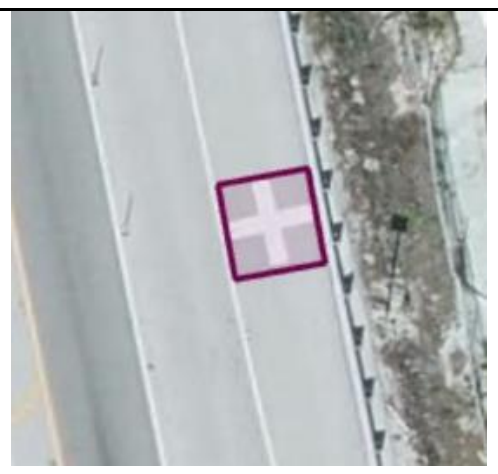 |

### *YOLOv3 – Turning Lane Detection Model*

You Only Look Once (YOLO) is mostly utilized for real-time object detection. Compared to other models such as R-CNN or Faster R-CNN, YOLO's standout advantage lies in its speed. Specifically, YOLO surpasses R-CNN and Faster R-CNN architectures by a factor of 1000 and 100, respectively (36). This notable speed advantage stems from the fact that while other models first classify potential regions and then identify items based on the classification probability of those regions, YOLO can predict based on the entire image context and perform the entire image analysis with just one network evaluation. Initially introduced in 2016 (37), YOLO has seen subsequent versions including YOLOv2 (38) and YOLOv3, which introduced improvements in multi-scale predictions. YOLOv4 (39) and YOLOv5 (40) were subsequently released in 2020 and 2022, respectively. For this study, YOLOv3 was chosen due to its suitability for the dataset at hand and its ease of application with available resources. Unlike other detection models, YOLOv3 employs Darknet-53 as its backbone, which is a deeper version of Darknet-19, the backbone of YOLOv2. Darknet-53 consists of 53 convolutional layers (41), enhancing both the accuracy and speed of the model, performing twice as fast as ResNet152 (36). **Figure 4** shows images generated at the conclusion of the detection layers in YOLOv3, illustrating the three-scale detection process. This process involves applying a 1x1 detection kernel on feature maps situated at three distinct areas and sizes within the network.

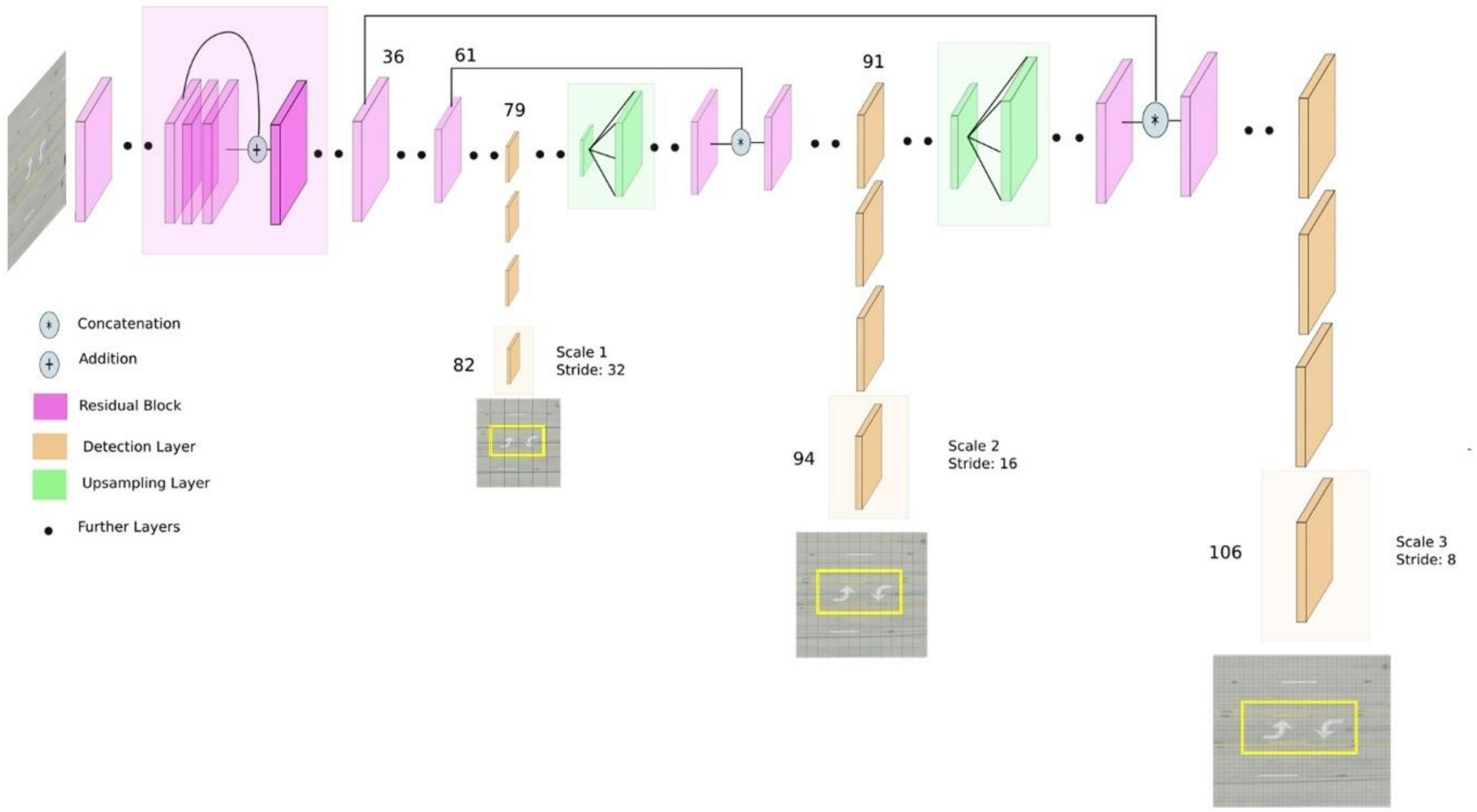

**Figure 4** YOLO v3 network architecture center lane example, adapted from (42).

### *Turning Lane Detector*

The training data prepared for the 12-class and 4-class turning lane models were manually labelled 23,669 and 8,241 features respectively, on the aerial images of Miami-Dade, Gulf, Santa Rosa, Hillsborough, and Broward counties using the Deep-Learning Toolbox in ArcGIS Pro. For the 12-class model, the "left_only" class had 3,021 features (12.3% of training data), "right_only" class had 2,179 features (8.9%), "left_through" class had 1,627 features (6.6%), "right_through" class had 1,583 features (6.4%), "through" class had 2,233 features (9.1%), "left_right_through" class had 920 features (3.7%), "bicycle" class had 3,230 features (13.1%), "center" class had 3,043 features (12.4%), "left_right" class had 159 features (0.6%), "merge" class had 2,262 features (9.2%), "u_turn" class had 632 features (2.6%), and "none" class had 2,780 features (11.3%). On the other hand, the 4-class model had 2,178 features (26.4%) for the "left_only" class, 2,060 features (25%) for the "right_only" class, 1,884 features (22.9%) for the "center" class, and 2,119 features (25.7%) for the "none" class. The "none" class labels for the 12-class model were made up of randomly selected roadway markings that describe all other features than the turning lane markings of interest including railroad crossings, stop, slow, only, and speed limit markings. While the "none" class labels for the 4-class model were made up of all other visible markings or features except "left_only", "right_only", and "center". Some features which were common for instance, left only, right only, and bicycle were more prevalent in the training data for the turning lane model with 12 classes while less common features like "left_right," "u_turn," and "left_right_through" had lower representation. To address this imbalance and enhance dataset diversity, some of these less common features were duplicated. However, to prevent potential issues like model overfitting and bias resulting from duplication, data augmentation techniques such as rotation were employed. Rotation involved randomly rotating the training data features at various angles to generate additional training data, thereby increasing dataset diversity. This approach enabled the model to recognize objects in different orientations and positions, which was particularly advantageous in this study, given that features like left-only and right-only lanes can appear in multiple travel directions at intersections. A 90 degrees rotation was applied to the training data. Finally, 468,176 exported image chips containing 567,876 features were used to train the 12-class model and 144,224 image chips containing 185,560 features were used to train the 4-class model. It is important to mention that a single image chip may contain multiple features.

Each class within the training data was distinctively labeled to ensure clarity. These classes distinctly categorized the turning lane features and non-features identified in the input image by the detector, thereby enhancing the model's detection accuracy. The output data was then sorted into categories such as left only, right only, center, and none, utilizing a class value field. The training dataset for the 12-class model comprised turning lane features from Miami-Dade (169,592), Hillsborough (286,584), Santa Rosa (4,708), and Gulf (8,556) counties, while the 4-class model training dataset included data from Broward (6,712), Miami-Dade (112,552), Hillsborough (53,136), Santa Rosa (5,324), and Gulf (7,836) counties. These counties were chosen for training to ensure a balanced model performance, considering variations in image resolutions and roadway infrastructure development, as well as the presence of diverse turning lane markings on the roadways.

The object detection model's adjustable parameters and hyperparameters encompass several factors such as the learning rate, input image dimensions, number of epochs, batch size, anchor box dimensions and ratios, as well as the percentages of training and testing data. Visualization of the machine learning model's evaluation metrics was presented through graphs depicting average precision, validation loss, and training loss (**Figure 5a** and **5b**). Validation loss and mean average precision were computed on the validation dataset, which constituted 30% of the input training dataset. Key parameters influencing object detection performance include the batch size, learning rate, and training epoch. The learning rate dictates the rate at which the model acquires new insights from training data, striking a balance between precision and convergence speed. Optimal learning rates of 1.096e-06 and 3.311e-06 were utilized to train the 12-class and 4-class models, respectively. Batch size refers to the number of training samples processed per iteration, with larger sizes facilitating parallel processing for faster training, albeit requiring more memory. In contrast, smaller batch sizes enhance model performance on new data by increasing randomness during training. Given the multi-class nature and high data complexity of the developed models, a batch size of 64 was adopted to enhance performance. The anchor box defines the size, shape, and positioning of the detected object, with 9 anchor boxes employed for model training. The epoch number denotes the iterations the model undergoes training, representing the number of times the training dataset is processed through the neural network. In this study, 70% of the training data was utilized for model training, selected randomly to ensure representativeness. A 30% (15%-15%) split of the training dataset was used for validating and testing the model's performance during training. In other words, 140,453 and 43,268 randomly selected image chips were used to assess the 12-class and 4-class models' accuracy respectively during the model training process. The determination of the training and test data split primarily hinged on the size of the training dataset. For datasets exceeding 10,000 samples, a validation size of 20-30% was deemed sufficient to provide randomly sampled data for evaluating the model's performance. A default 50% overlap between the label and detection bounding boxes was considered a valid prediction criterion. Subsequently, recall and precision were computed to assess the true prediction rate among the original labels and all other predictions, respectively. Unlike previous versions, YOLOv3 calculates the likelihood of an input belonging to a specific label using individual logistic classifiers instead of the SoftMax function (37, 38). During model training, binary cross-entropy loss was employed for each label to compute the classification loss, rather than mean square error. Consequently, logistic regression was utilized to predict both object confidence and class predictions, streamlining computational complexity and enhancing model performance (36).

The train-validation loss graphs for the 12-class and 4-class models are depicted in Figures 5a and 5b, respectively. These graphs illustrate the evaluation of the model's performance concerning new and training data. A higher loss value suggests more errors in the model, while a lower loss value indicates fewer errors. Analysis of the graphs reveals that the 4-class model exhibited a larger difference between training and validation losses, with the training loss considerably lower than the validation loss (**Figure 5b**). Conversely, the 12-class model displayed minimal disparity between training and validation loss values (**Figure 5a**), indicating that both models fit equally well with new data, albeit the 4-class model exhibiting better fitting with the training data and fewer errors compared to the 12-class model. The average accuracy achieved by the developed 12-class model was 0.84, while the 4-class model averaged 0.85. Consequently, it can be inferred that the developed detector performs quite well.

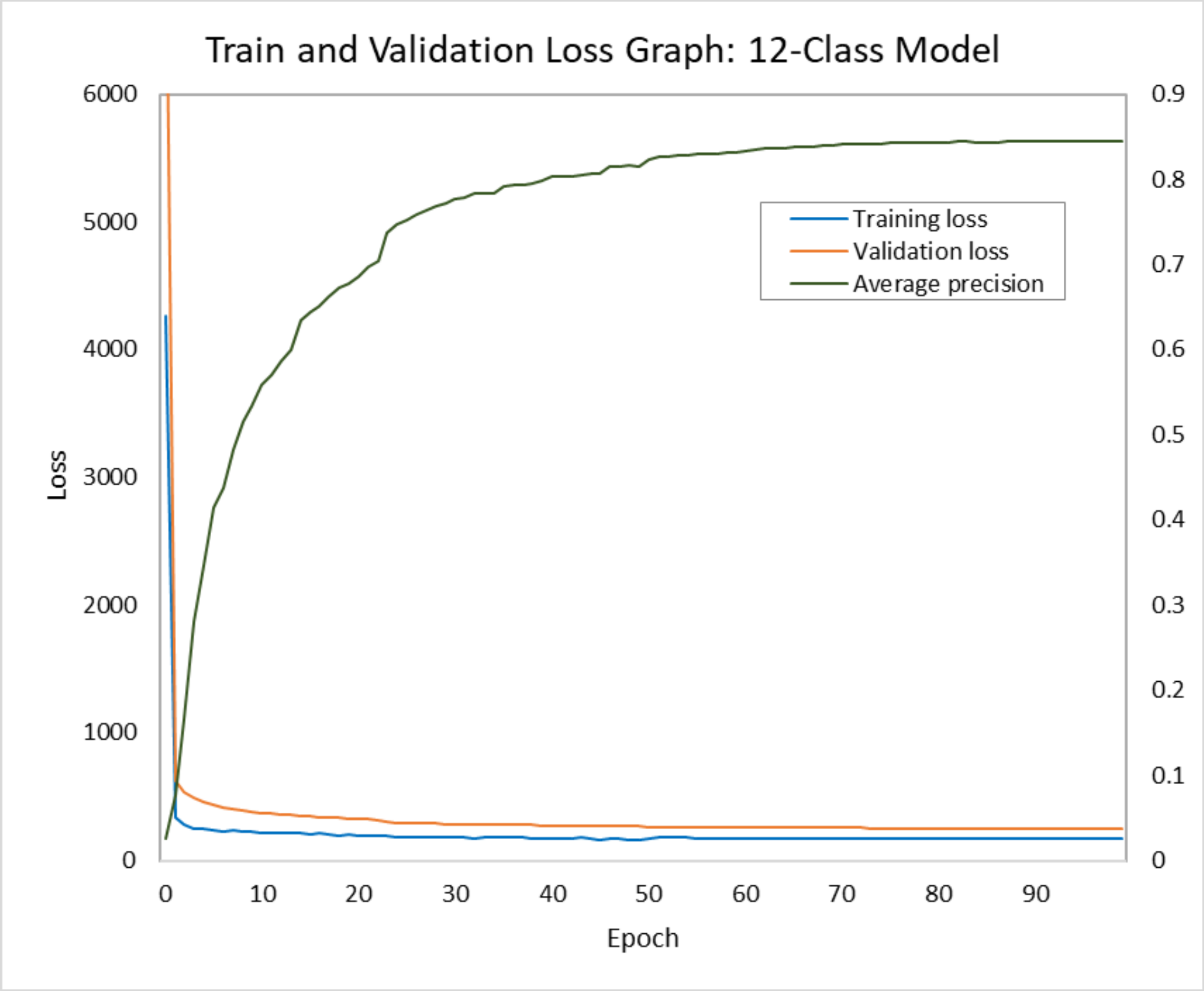

**(5a)**

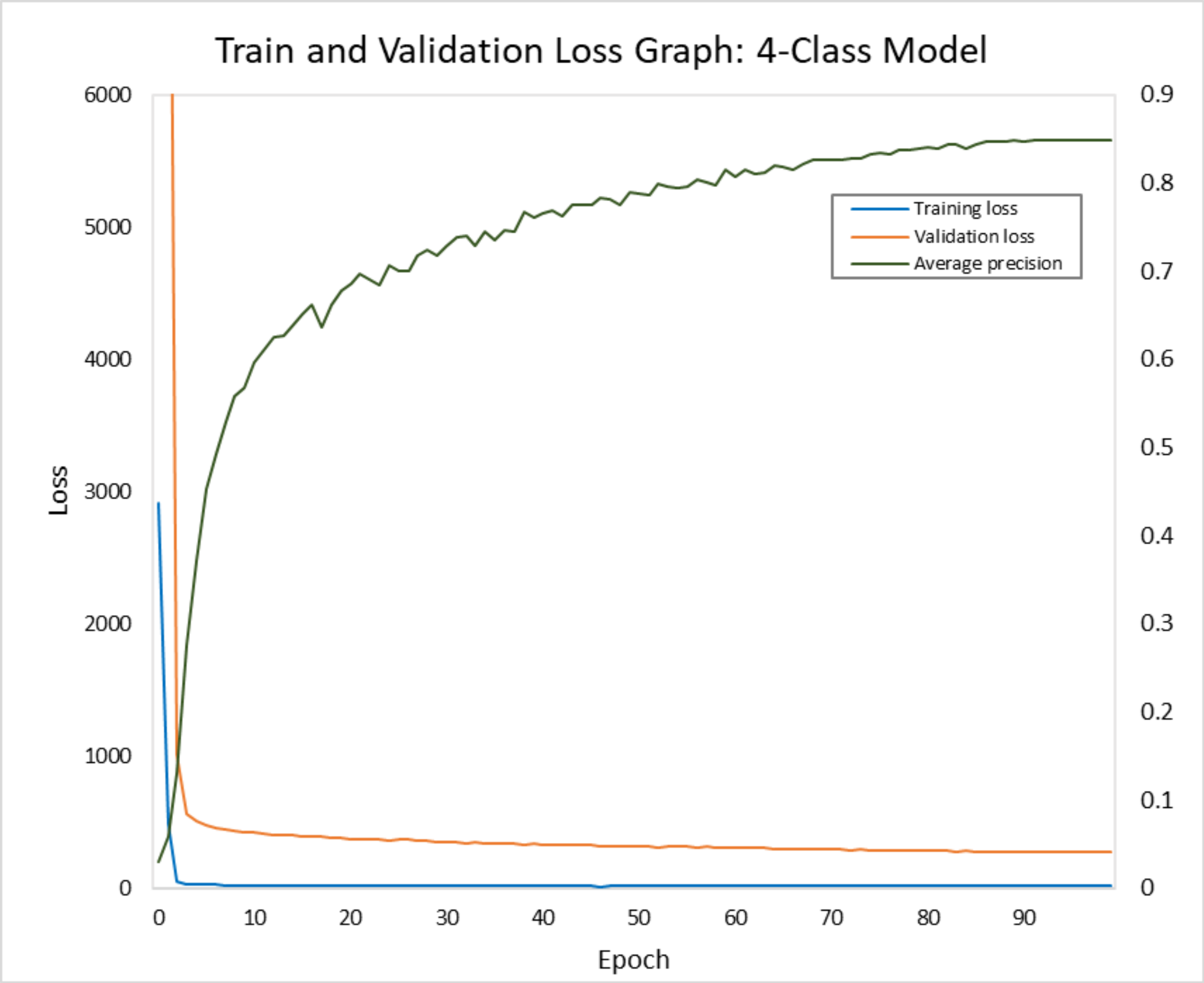

**(5b)**

**Figure 5:** Developed YOLOv3 turning lane model train and validation loss graph (a) 12-class model, and (b) 4-class model.

***Mapping Turning Lanes***

The turning lane detector was initially evaluated on individual photos. **Figure 6b** demonstrates how the detector correctly outlines turning lanes with bounding boxes using the detection's confidence score. Due to the complexity of multi class models, a threshold of 0.05 was employed to capture all features detected with low confidence levels. It is important to note that lower detection thresholds increase false positives which results in lower precision, increase computational workload which intend increases detection time, and increases detection of irrelevant features or noise. However, lower detection thresholds increase model's sensitivity to detecting faint or partially visible features, while increasing recall. With higher recall, the model is more likely to identify and detect all instances of the target object class and reduce the chances of missing any objects. More than 10% overlap between two bounding boxes was avoided to minimize duplicate detections. To lessen information loss from the margins of the detection images, a padding parameter of fifty-six was added to the boundary pixels on the outside of the image. The detector was trained on 256 x 256 sub-images with a stride of 128 by 128 pixels, yielding a 50% overlap with the following image chip, and a resolution of 0.5 feet per pixel. It should be noted that utilizing huge photos with object detection techniques is impractical since the cost of computing grows rapidly.

The mapping procedure was conducted at the county level because the detector performs very well on single photos. **Figure 6a** provides a summary of this procedure. The photographs in the folder labeled "masked images" were first picked out and iterated through the detector. An output file of all the identified turning lanes in that county was created once all photos had been sent to the detector. Confidence scores were included in the output file. This file was used to map turning lanes. Note that the model can detect turning lane markings from images with a resolution ranging from 1.5 ft down to 0.25 ft or higher. However, the model has not been tried on any images with resolution lower than the ones provided by the Florida APLUS system. From the observations, the model made some false detections in a few instances. These were outlined and discussed in the results. **Figure 6b** shows some examples of the detected features and some observed false detections or misclassification.

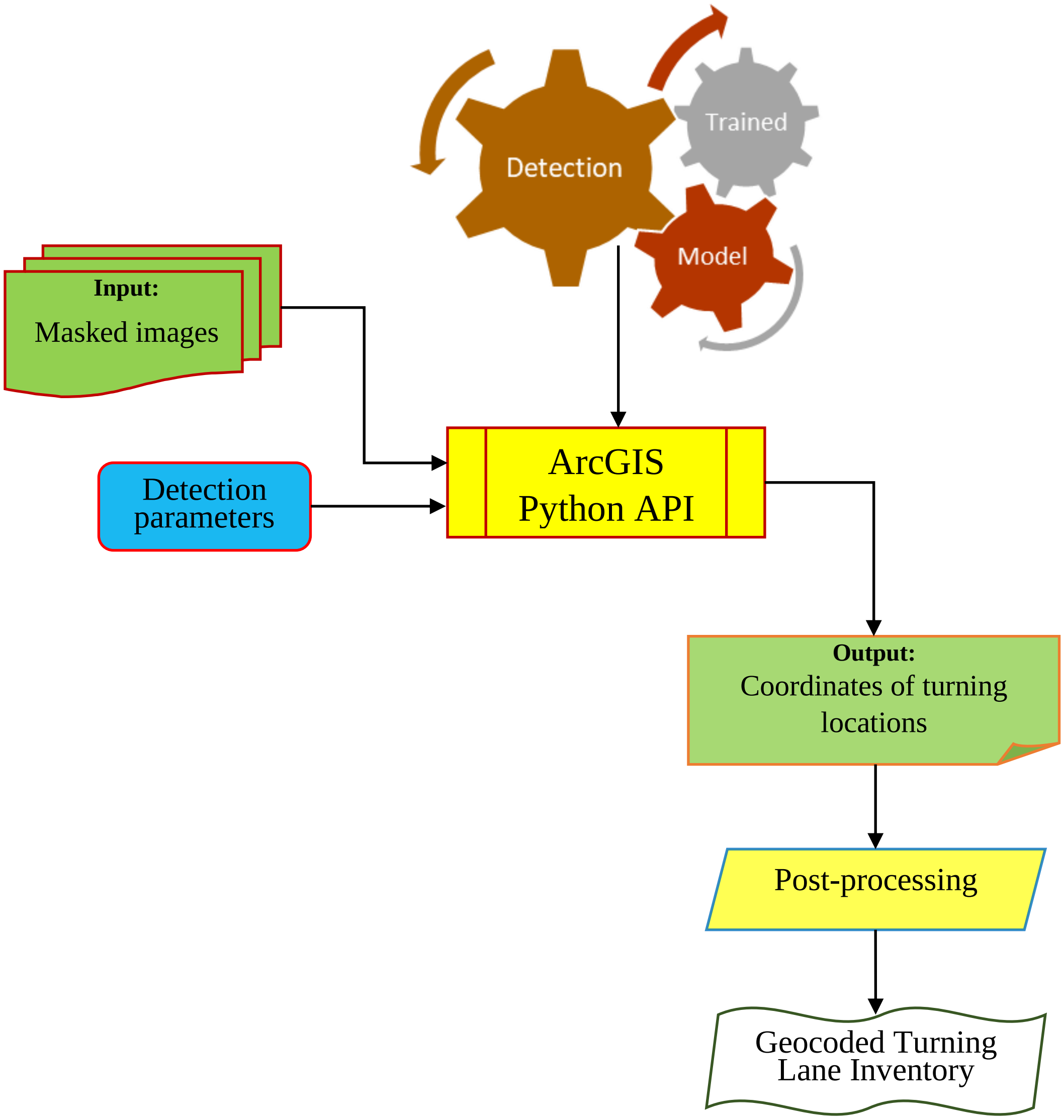

**(6a)**

Name Examples of turning lane feature detection from images

Multiple features

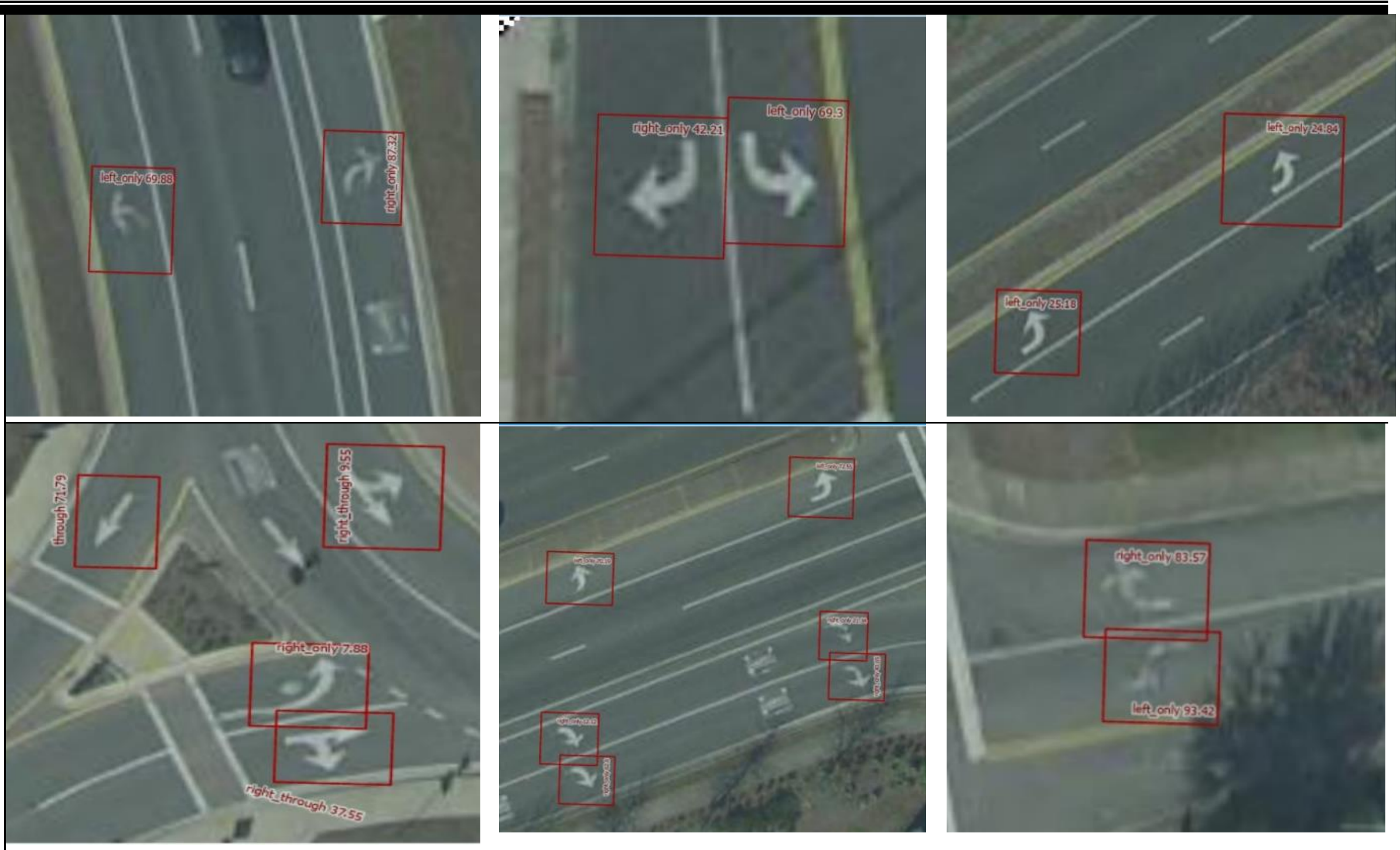

Left, Right, and Center

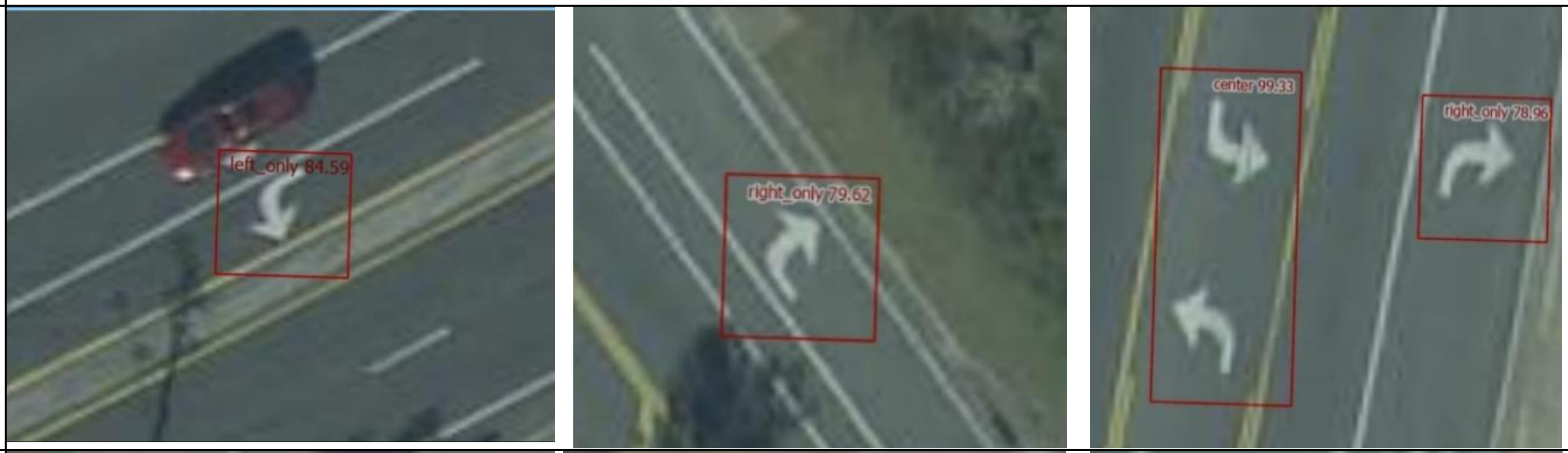

Left_through, right_through, and through

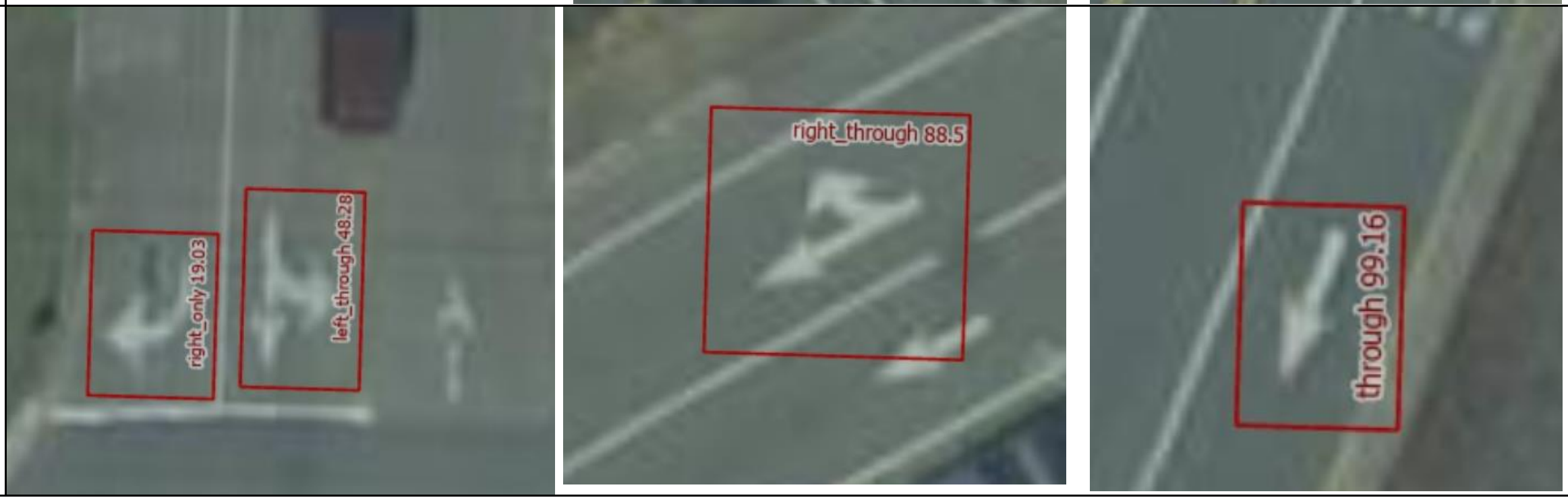

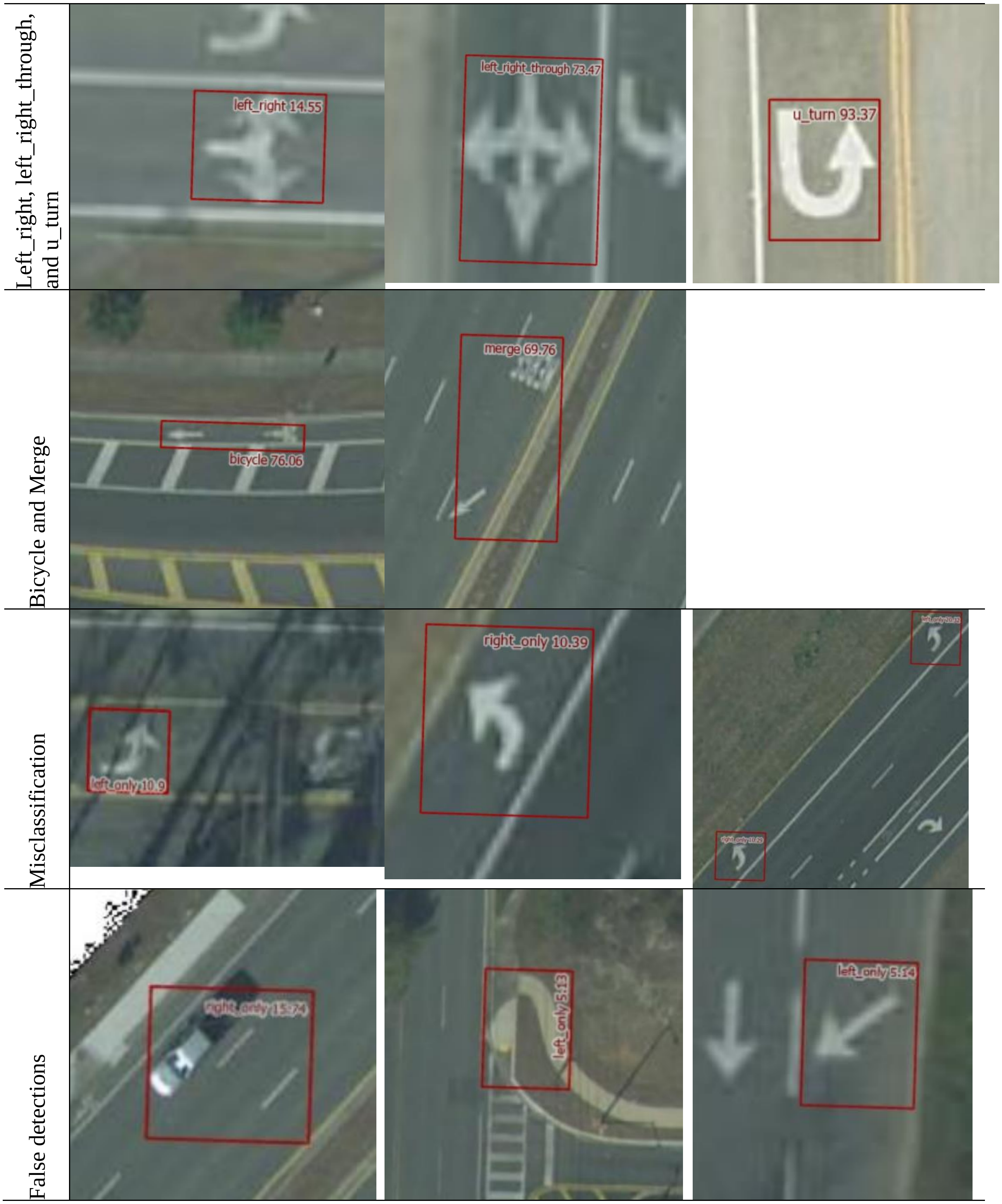

**(6b)**

**Figure 6** (a) Turning lane detection framework, and (b) turning lane detection polygons and confidence scores on images from Leon County.

### *Post-processing*

After applying the model on the aerial obtained aerial images in Leon County, the total number of observed left, right, and center detections in Leon County from the 12 class model and 4-class model were 2,736 and 3,165 respectively. The redundant detections caused by the overlapping distance on images are removed at the post-processing stage. For various analytical objectives, turning configurations of state and local roadways can also be divided into categories. Non-maximum suppression was used to filter the detected turning configurations by selecting and keeping detections that overlap and have the highest confidence level. Detected turning lane markings with more than a 10% overlap and lower confidence levels were removed. Detected features were also converted from polygon shapefiles into point shapefiles for analysis.

## RESULTS AND DISCUSSIONS

### *Case Study: Overall Performance Evaluation Using the Ground Truth Data*

#### *Experimental Design*

In the first case study, the model's performance is assessed together with its accuracy and completeness, and the results are contrasted with training data. Leon County was utilized as the county where ground truth data were collected and the developed models' performances were assessed together with their accuracies and completeness, and the results were contrasted with training data. A full, manually positioned Ground Truth (GT) dataset was created in Leon County as a proof of concept. After visual inspection, a total of 2,566 visible turning lane markings for left, right, and center lanes turning lane could encompass several consecutive features that collectively define that lane. GT data was collected using the masked photographs as the background. Note that each turning lane may exhibit a range of 1-6 features, such as left-only or right-only indicators. Consequently, a single lane contains multiple turning lanes aligned in parallel, each lane may consist of several features. Within this framework, any feature that is inaccurately classified amidst correctly classified features within the same lane is deemed a false positive or a misclassified turning lane. For instance, in a left-only lane with four features, if three are correctly identified as left, and the fourth is misclassified as right or center, the entire lane is considered a false positive or misclassified detection. **Figure 7** shows the GT dataset and detected turning lane markings in Leon County. Although some of the turning lane markings, especially the left only marking, were missed by the model, the overall performance of the model was reliable. This is mainly because of reasons such as occlusions, faded markings, shadows, poor image resolution, and variety of the pavement marking design.

As noted, the suggested model has identified turning lane markers with a minimum confidence score of 5%. For the purposes of this case study, the model identified turning lane markers (M) on both state and local roads in Leon County, which were then retrieved. On the GT, a similar location-based selection methodology was used. The suggested model's performance was assessed by examining the points that were discovered within the polygons and vice versa at various confidence levels of 75%, 50%, 25%, 10%, and 5% because the GT and M are points and polygons, respectively (**Table 2**). The performance of the 2 developed models were assessed.

The suggested models' performance was assessed using the criteria of completeness (precision), correctness (recall), and quality (intersection over union) and visualized using a circus plot (43) in **Figures 8a and 8b**. These criteria were initially utilized in (44) and (45) for the goal of highway extraction, and they are now often used for performance evaluation of the related models (46, 47). The following selection criteria are necessary to determine the performance evaluation metrics:

i. <u>GT</u>: Number of GT turning lane points, and
ii. <u>M</u>: Number of Model detected turning lane markings

iii. False Negative (FN): # of GT turning lane points not found within M turning lane polygon,
iv. False Positive (FP): # of M turning lane polygons with no GT turning lane point,
v. True Positive (TP): # of M turning lane polygons with GT turning lane point,

Performance evaluation metrics:

Completeness $=\frac{GT-FN}{GT} * 100\%$, true detection rate among GT turning lane (recall)

Correctness $=\frac{M-FP}{M} * 100\%$, True detection rate among M turning lane (precision)

Quality $=\frac{GT-FN}{GT+FP} * 100\%$, True detection among M turning lane plus the undetected GT turning lane (Intersection over Union: IoU)

This case study's major goal is to assess the accuracy and performance of the proposed model's predictions and contrast them with a ground truth dataset. Separate evaluation analysis will be performed using each “left_only”, “right_only”, and “center” detections of the developed models. The accuracy and performance on turning lane model will be assessed after evaluating the model's correctness (precision) and completeness (recall) using a complete ground truth dataset and measuring the f-1 score. The f-1 score calculates the harmonic mean using the precision and recall values. It is the appropriate evaluation metric when dealing with imbalanced datasets. It is crucial in object detection tasks where missing actual objects is more detrimental than incorrectly classifying background regions as objects. The f1-scores of the developed models were compared and visualized using a circus plot in **Figure 9**.

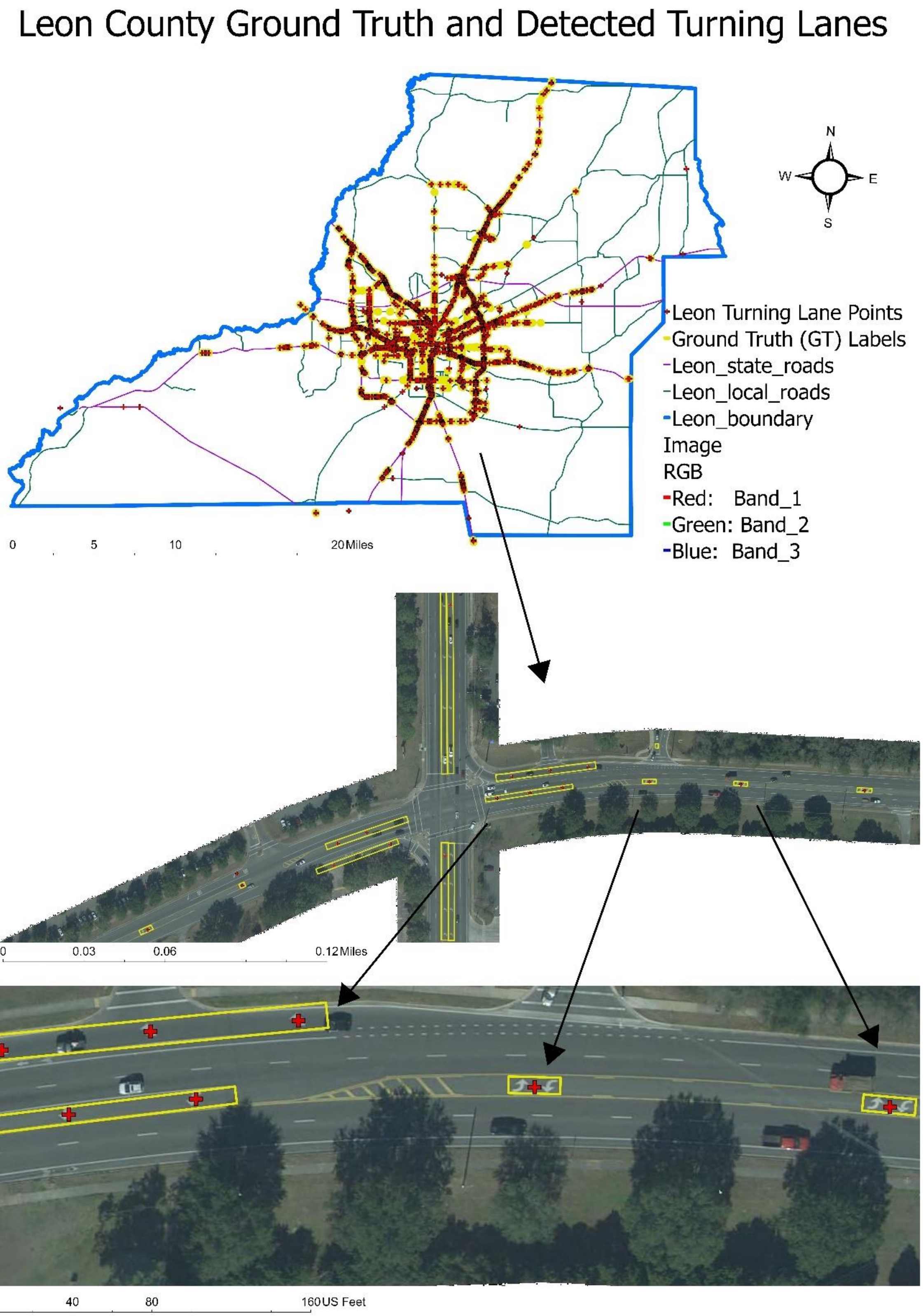

**Figure 7** Manually labeled Ground Truth turning lane markings (GT) and detected turning lane markings in Leon County.

Based on the findings, we observe that this automated 12-class turning lane detection and mapping model can averagely detect and map 23% of the turning lanes with 94% precision at 75% confidence level. At a lower confidence level of 5%, it can detect 73% of the turning lanes with 96% precision. At 75% confidence level, the average quality of the model's detection is 16% whereas the average quality of the model is 59% at 5% confidence level. On average, the 4-class turning lane model recorded 42% of detections with 99% precision at 75% confidence level. At a lower 5% confidence level, 80% of detections were observed at 97% precision. The average quality of the 4-class model detections at 75% confidence level is 28% and 68% at 5% confidence level. Higher accuracy was achieved at low confidence levels since there is a higher recall and more room is given to increase the number of detections. That is, from the observations, detected turning lane markings that had occlusions from vehicle or trees, shadows, and faded markings had lower confidence levels. Therefore, reducing the confidence level threshold adds these detected features to the total number of detections. The new detections allowed into the pool for valuation includes more true positives, less false positives and less or zero false negatives.

With the increase in the number of true positives as confidence decreases, the accuracy of the model increases since the accuracy is described based on the relationship between the number of true positives, total number of detections, the high resolution aerial images and the false negatives. It can be observed that the 4-class turning lane detection model recorded higher accuracies than the 12-class model. This is due to the less complexity or lower number of classes of the 4-class model which increases model's performance. The summarized model performance evaluation in **Table 2** and visualized in **Figures 8a and 8b**. It can also be observed that the poor distinctiveness of the detection features affected detection performance. As stated earlier, the observed difference between the left only and right only turning markings, which is just a lateral inversion of the other marking, made it less unique and therefore resulted in low detection performance. When the left turn is flipped horizontally, it becomes a right turn and vice versa. On the other hand, center lane which was trained using a distinct shape recorded better detection results than the left only and right only lanes.

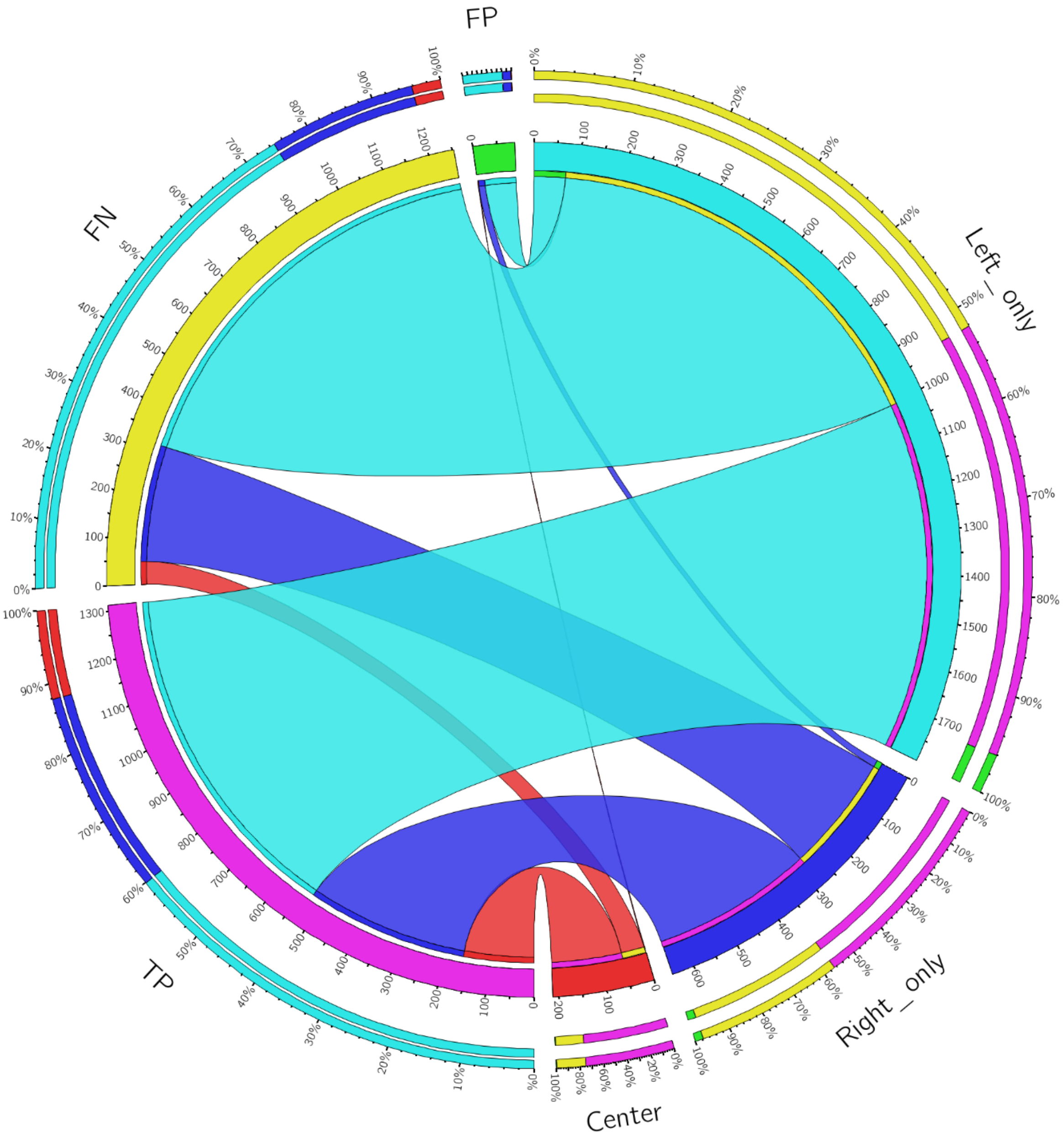

**(8a)**

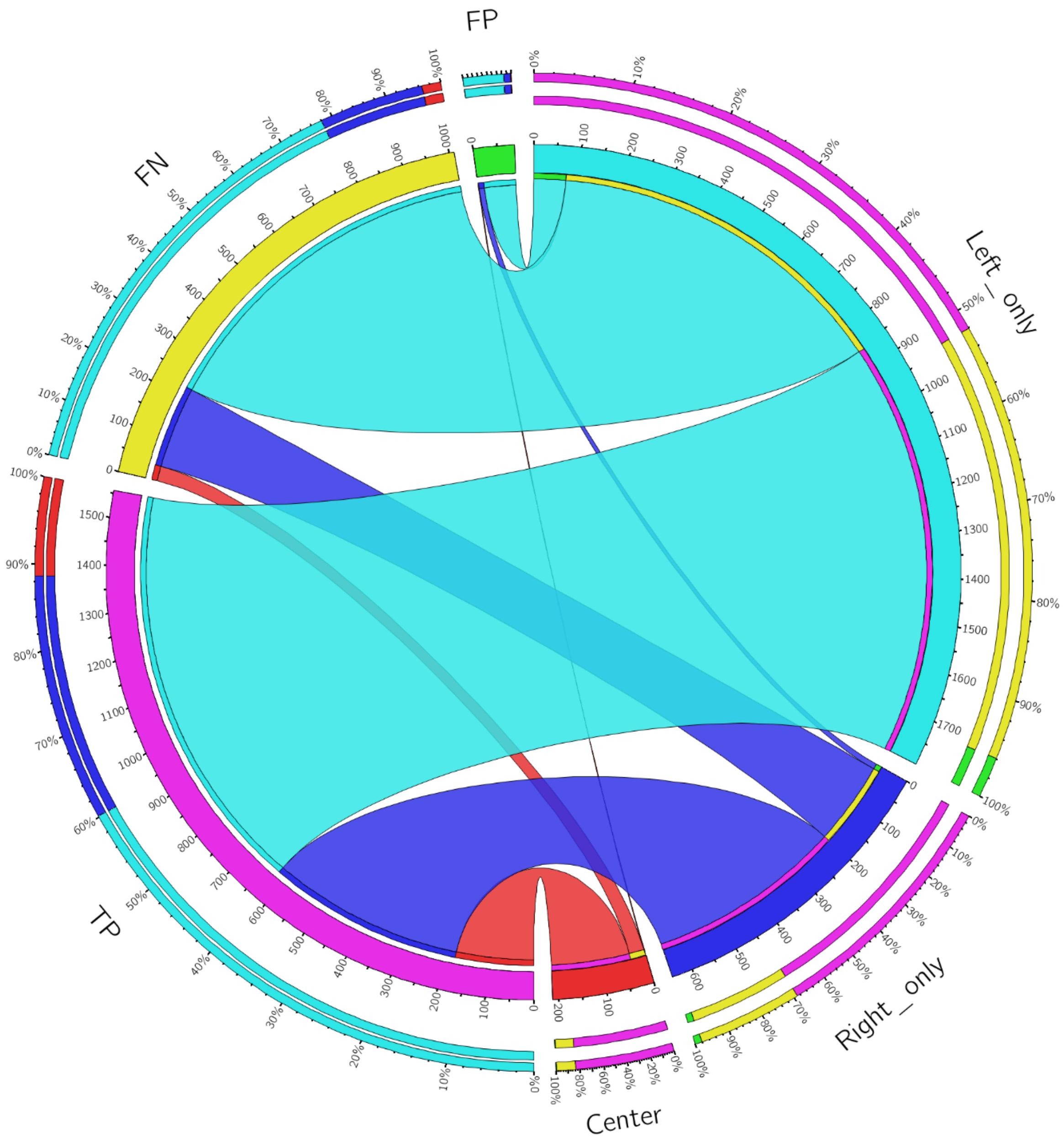

**(8b)**

**Figure 8** Visualization (circus plot) of performance evaluation metrics between the ground truth (GT) and predictions made by the YOLOv3 based (**a**) 12-class turning lane model and (**b**) 4-class turning lane model for detecting left_only (turquoise), right_only (blue), and center (red). The circus plot also shows the distribution of the true positives (magenta), false negatives (yellow), and false positives (green). The links between the classes show the number of true positives (correctly classified), false negatives (unclassified), and false positives (misclassified) in each class; the thickness of the links describe their percentages. The size of the radii of the inner segments depicts the total value of the fields in ascending order. The outer concentric bars depict the percentages of the values in descending order. From **8b** (4-class model), about 70% of right only detections are true positives while only 56% were true positives in **8a** (12-class model).

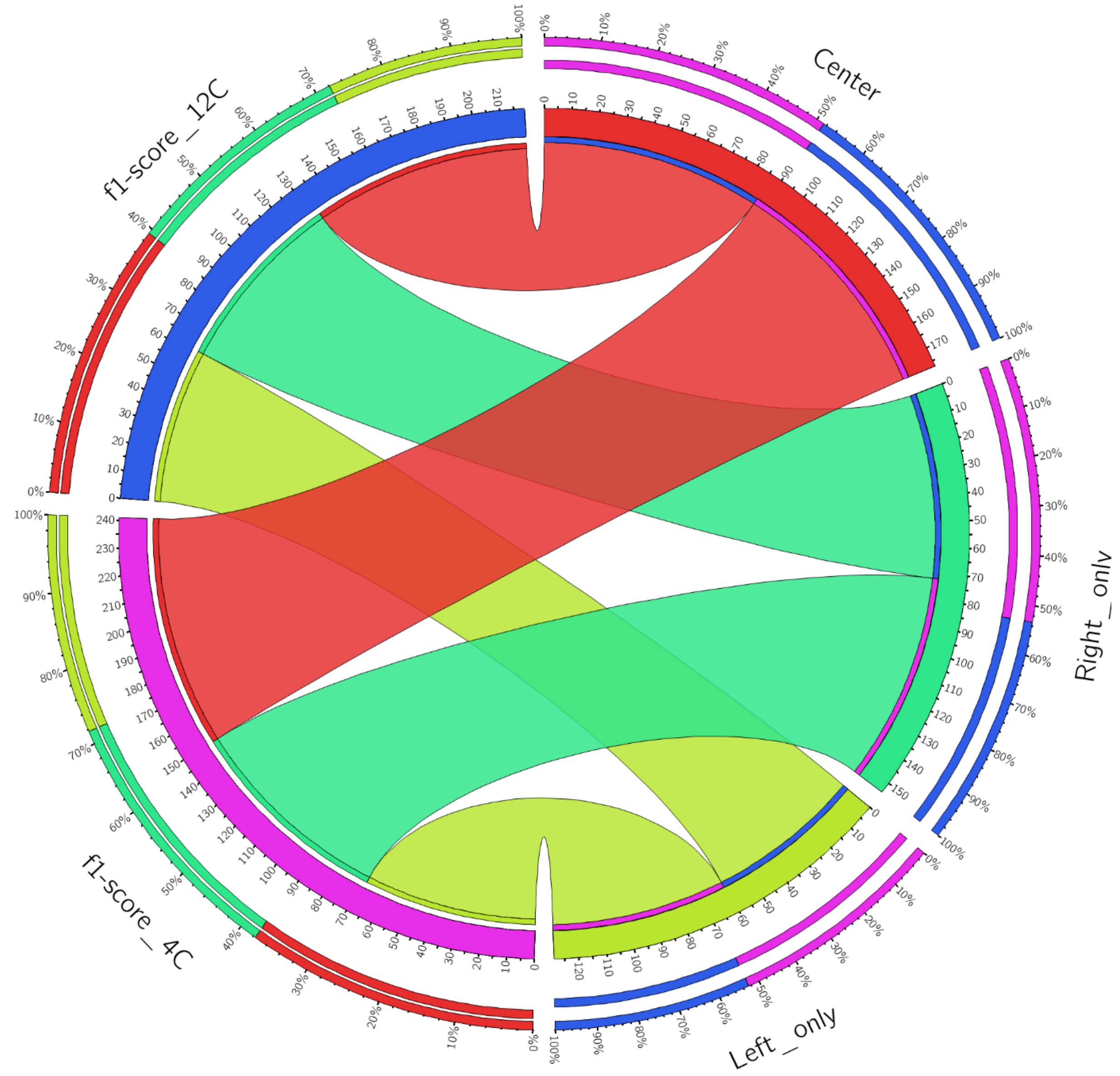

**Figure 9** Visualization (circus plot) of f1-score comparison between the 12-class and 4-class YOLOv3 based turning lane model for detecting left_only (green yellow), right_only (shamrock), and center (red). The circus plot also shows the distribution of the 4-class model's f1-score (magenta) and the 12-class model's f1-score (blue). The links between the classes show the f1-scores in each class; the thickness of the links describe their percentages. The size of the radii of the inner segments depicts the total value of the fields in ascending order. The outer concentric bars depict the percentages of the values in descending order. The f1-score of the 4-class model is more than 50% in all classes, indicating a better performance than the 12-class model. Also, the total value of the 4-class model's f1-score is ~240 which is higher than the 12-class model which is ~215.

### *Detecting Turning Lanes in Duval County*

Additionally, the model was used to detect turning lanes in Duval County. The detected turning lanes were classified under different confidence levels. The final list is shown in **Table 2**. The detected turning lanes have been visualized in **Figure 10**. The extracted road geometry data can be integrated with crash and traffic data especially at intersections to advise policy makers and roadway users. That is, they can be used for a variety of purposes such as identifying those markings that are old and invisible, comparing the turning lane locations with other geometric features like crosswalks, school zones and analyzing the crashes occurring around the turning lanes at intersections.

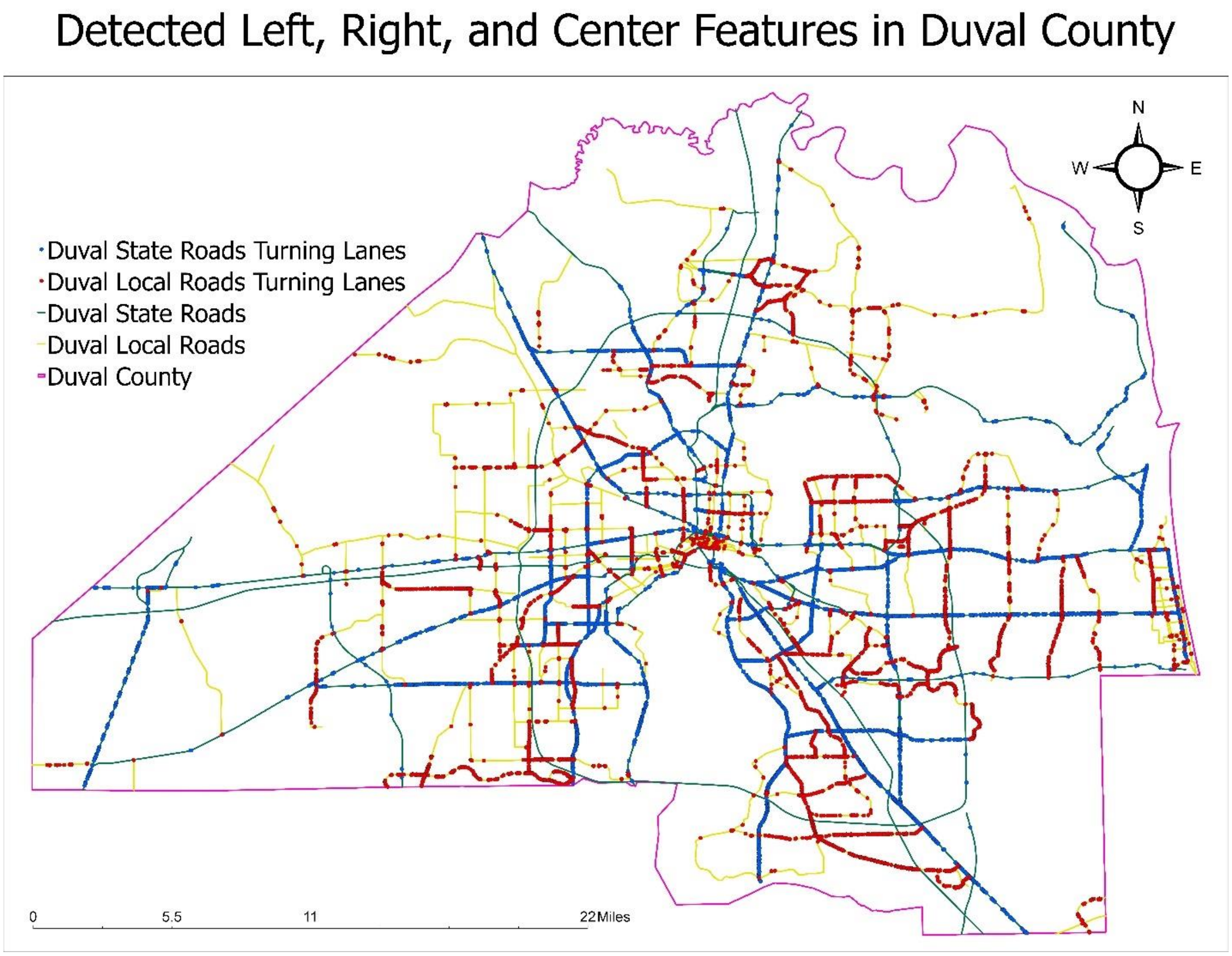

**Figure 10** Model-detected turning lanes in Duval County

**Table 2** Model performance evaluations

| Leon County Ground Truth Comparison Analysis | | | | | | | | |
|---|---|---|---|---|---|---|---|---|
| **4-class model** | | | | | | | | |
| **left_only: GT = 1723** | | | | | | | | |
| **Confidence (%)** | **M** | **TP** | **FP** | **FN** | **Completeness (%)** | **Correctness (%)** | **Quality (%)** | **F1-score (%)** |
| 75 | 283 | 275 | 8 | 1448 | 15.96 | 97.17 | 15.89 | 27.42 |
| 50 | 457 | 436 | 21 | 1287 | 25.30 | 95.40 | 25.00 | 40.00 |
| 25 | 666 | 626 | 40 | 1097 | 36.33 | 93.99 | 35.51 | 52.41 |
| 10 | 821 | 764 | 57 | 959 | 44.34 | 93.06 | 42.92 | 60.06 |
| 5 | 1008 | 935 | 73 | 788 | 54.27 | 92.76 | 52.06 | 68.47 |
| **right_only: GT = 632** | | | | | | | | |
| **Confidence (%)** | **M** | **TP** | **FP** | **FN** | **Completeness (%)** | **Correctness (%)** | **Quality (%)** | **F1-score (%)** |
| 75 | 129 | 127 | 2 | 505 | 20.09 | 98.45 | 20.03 | 33.38 |
| 50 | 216 | 211 | 5 | 421 | 33.39 | 97.69 | 33.12 | 49.76 |
| 25 | 341 | 334 | 7 | 298 | 52.85 | 97.95 | 52.27 | 68.65 |
| 10 | 400 | 390 | 10 | 242 | 61.71 | 97.50 | 60.75 | 75.58 |
| 5 | 456 | 444 | 12 | 188 | 70.25 | 97.37 | 68.94 | 81.62 |
| **center: GT = 211** | | | | | | | | |
| **Confidence (%)** | **M** | **TP** | **FP** | **FN** | **Completeness (%)** | **Correctness (%)** | **Quality (%)** | **F1-score (%)** |
| 75 | 104 | 104 | 0 | 107 | 49.29 | 100.00 | 49.29 | 66.03 |
| 50 | 133 | 133 | 0 | 78 | 63.03 | 100.00 | 63.03 | 77.33 |
| 25 | 157 | 157 | 0 | 54 | 74.41 | 100.00 | 74.41 | 85.33 |
| 10 | 166 | 166 | 0 | 45 | 78.67 | 100.00 | 78.67 | 88.06 |
| 5 | 178 | 177 | 1 | 34 | 83.89 | 99.44 | 83.49 | 91.00 |
| **12-class model** | | | | | | | | |
| **left_only: GT = 1723** | | | | | | | | |
| **Confidence (%)** | **M** | **TP** | **FP** | **FN** | **Completeness (%)** | **Correctness (%)** | **Quality (%)** | **F1-score (%)** |
| 75 | 11 | 10 | 1 | 1713 | 0.58 | 90.91 | 0.58 | 1.15 |
| 50 | 62 | 57 | 5 | 1666 | 3.31 | 91.94 | 3.30 | 6.39 |
| 25 | 271 | 255 | 16 | 1468 | 14.80 | 94.10 | 14.66 | 25.58 |
| 10 | 553 | 513 | 40 | 1210 | 29.77 | 92.77 | 29.10 | 45.08 |
| 5 | 858 | 787 | 71 | 936 | 45.68 | 91.72 | 43.87 | 60.98 |
| **right_only: GT = 632** | | | | | | | | |
| **Confidence (%)** | **M** | **TP** | **FP** | **FN** | **Completeness (%)** | **Correctness (%)** | **Quality (%)** | **F1-score (%)** |
| 75 | 11 | 10 | 1 | 622 | 1.58 | 90.91 | 1.58 | 3.11 |
| 50 | 62 | 60 | 2 | 572 | 9.49 | 96.77 | 9.46 | 17.29 |
| 25 | 182 | 176 | 6 | 456 | 27.85 | 96.70 | 27.59 | 43.24 |
| 10 | 304 | 291 | 13 | 341 | 46.04 | 95.72 | 45.12 | 62.18 |
| 5 | 382 | 367 | 15 | 265 | 58.07 | 96.07 | 56.72 | 72.39 |
| **center: GT = 211** | | | | | | | | |
| **Confidence (%)** | **M** | **TP** | **FP** | **FN** | **Completeness (%)** | **Correctness (%)** | **Quality (%)** | **F1-score (%)** |
| 75 | 98 | 98 | 0 | 113 | 46.45 | 100.00 | 46.45 | 63.43 |
| 50 | 129 | 129 | 0 | 82 | 61.14 | 100.00 | 61.14 | 75.88 |
| 25 | 148 | 148 | 0 | 63 | 70.14 | 100.00 | 70.14 | 82.45 |
| 10 | 155 | 154 | 1 | 57 | 72.99 | 99.35 | 72.64 | 84.15 |
| 5 | 160 | 159 | 1 | 52 | 75.36 | 99.38 | 75.00 | 85.71 |
| | | | | | | | | |
| **Detection of Right, Left, and Center Markings in Duval County** | | | | | | | | |
| **Left** | | | | | | | | |

| Confidence Level (%) | Detected Turning Lanes | | |
|---|---|---|---|
| | State Roads | Local Roads | Total |
| 75 | 1238 | 971 | 2209 |
| 50 | 2167 | 1722 | 3889 |
| 25 | 3607 | 2819 | 6426 |
| 10 | 4315 | 3346 | 7661 |
| 5 | 4969 | 3768 | 8737 |

**Right**

| Confidence Level (%) | Detected Turning Lanes | | |
|---|---|---|---|
| | State Roads | Local Roads | Total |
| 75 | 459 | 311 | 770 |
| 50 | 822 | 607 | 1429 |
| 25 | 1381 | 1066 | 2447 |
| 10 | 1661 | 1347 | 3008 |
| 5 | 1946 | 1631 | 3577 |

**Center**

| Confidence Level (%) | Detected Turning Lanes | | |
|---|---|---|---|
| | State Roads | Local Roads | Total |
| 75 | 317 | 406 | 723 |
| 50 | 430 | 533 | 963 |
| 25 | 478 | 627 | 1105 |
| 10 | 522 | 706 | 1228 |
| 5 | 634 | 886 | 1520 |

# CONCLUSIONS AND FUTURE WORK

This study explores the application of computer vision techniques for extracting roadway geometry, with a specific focus on Florida turning lanes as a pilot project. This innovative approach aims to replace labor-intensive manual inventory processes with computer vision technology, potentially reducing errors associated with manual data entry. By leveraging high-quality imagery, the developed system can accurately identify turning lane markings from images, offering a cost-effective alternative to human inventory development procedures. This advancement not only streamlines data collection but also enhances the quality of highway geometry data by minimizing manual errors.

The benefits of this automated roadway data extraction approach for transportation agencies are manifold. It facilitates the identification of outdated or obscured markings, enables the comparison of turning lane locations with other geometric features such as crosswalks and school zones, and supports the analysis of accidents occurring in proximity to these locations. However, the study also identifies notable limitations and offers recommendations for future research. Challenges arise from aerial images of roadways obstructed by tree canopies, limiting the identification of turning lane markings. Moving forward, the developed model can be integrated into roadway geometry inventory datasets, such as those used in the Highway Safety Manual (HSM) and the Model Inventory of Roadway Elements (MIRE), to identify and rectify outdated or missing lane markings.

Future research endeavors will focus on refining and expanding the capabilities of the model to detect and extract additional roadway geometric features. Additionally, there are plans to integrate the extracted left-only, right-only, and center lanes with crash data, traffic data, and demographic information for a more comprehensive analysis.

## ACKNOWLEDGEMENTS

This study was sponsored by the State of Florida Department of Transportation (DOT) grant BED30-977-02. The contents of this paper and discussion represent the authors' opinions and do not reflect the official views of the Florida Department of Transportation.

## AUTHOR CONTRIBUTIONS

The following authors confirm contribution to the paper with regards to Study conception and design: Richard B. Antwi, Samuel Takyi, Kimollo Michael, Alican Karaer, Eren Erman Ozguven, Ren Moses, Maxim A. Dulebenets, and Thobias Sando; Data collection: Richard B. Antwi, Samuel Takyi, Kimollo Michael, Eren Erman Ozguven, Ren Moses, Maxim A. Dulebenets, and Thobias Sando; Analysis and interpretation of results; Manuscript preparation: Richard B. Antwi, Eren Erman Ozguven, Ren Moses, Maxim A. Dulebenets, and Thobias Sando. All authors reviewed the results and approved the final version of the manuscript.